\documentclass{wlscirep}

\usepackage{multirow}
\usepackage{babelbib}

\title{SimMAT: Exploring Transferability from Vision Foundation Models to Any Image Modality}

\author[1,2,*]{Chenyang Lei}
\author[1,3,*]{Liyi Chen}
\author[4]{Jun Cen}
\author[1,3]{Xiao Chen}
\author[1,5]{Zhen Lei}
\author[2]{Felix Heide}
\author[6]{Ziwei Liu}
\author[4]{Qifeng Chen}
\author[1,5$\dagger$]{Zhaoxiang Zhang}

\affil[1]{ Center for Artificial Intelligence and Robotics, HKISI, CAS}
\affil[2]{Princeton University, Department of Computer Science}
\affil[3]{The Hong Kong Polytechnic University}
\affil[4]{The Hong Kong University of Science and Technology}
\affil[5]{State Key Laboratory of Multimodal Artificial Intelligence Systems, CASIA}
\affil[6]{Nanyang Technological University}
\affil[$\dagger$]{corresponding author. E-mail: zhaoxiang.zhang@ia.ac.cn}

\affil[*]{these authors contributed equally to this work}

\begin{abstract}
\newcommand{\exptrainingsctrach}{22.15}
\newcommand{\expsimmat}{53.88}

\newcommand{\mattrainingsctrach}{25.43}
\newcommand{\matrandomly}{58.89}
\newcommand{\matlinear}{63.96}
\newcommand{\mattranspose}{70.90}
\newcommand{\matours}{72.69}

Foundation models like ChatGPT and Sora that are trained on a huge scale of data have made a revolutionary social impact. However, it is extremely challenging for sensors in many different fields to collect similar scales of natural images to train strong foundation models. To this end, this work presents a simple and effective 
framework \texttt{SimMAT} to study an open problem: the transferability from vision foundation models trained on natural RGB images to other image modalities of different physical properties (e.g., polarization). SimMAT consists of a modality-agnostic transfer layer (MAT) and a pretrained foundation model. We apply SimMAT to a representative vision foundation model Segment Anything Model (SAM) to support any evaluated new image modality. Given the absence of relevant benchmarks, we construct a new benchmark to evaluate the transfer learning performance. Our experiments confirm the intriguing potential of transferring vision foundation models in enhancing other sensors' performance. Specifically, SimMAT can improve the segmentation performance (mIoU) from \exptrainingsctrach\% to \expsimmat\% on average for evaluated modalities and consistently outperforms other baselines. We hope that SimMAT can raise awareness of cross-modal transfer learning and benefit various fields for better results with vision foundation models. 
\end{abstract}
\begin{document}

\flushbottom
\maketitle
\thispagestyle{empty}

\section*{Introduction}

Foundation models have revolutionized computer vision~\cite{kirillov2023segment, bai2024sequential,bommasani2021opportunities} and natural language processing~\cite{kenton2019bert,brown2020language} across the fields, from personal assistance to self-driving vehicles and medical diagnosis~\cite{ma2024segment,zhou2024pre,yang2024vision,gehrig2024low,yako2023video,merchant2023scaling}. Diverse downstream tasks rely directly or indirectly on foundation models by finetuning foundation models that are pretrained on large-scale data with pretext tasks~\cite{pan2009survey}. However, while diverse types of sensors~\cite{wu2023medical,huang2023polarization,lei2022shape,lei2020polarized,sun2022seeing,gallego2020event,dong2016hyperspectral,tseng2021neural} are applied in various domains in the world, e.g., medical imaging, robotics, and fundamental science, not all of them benefit from the development of foundation models. 
This is because it is challenging for other sensors~\cite{wang2024sub,mao2024multimodal} to collect large-scale training data like natural images, as shown in Figure~\ref{fig:framework}. 

This work explores the following problem: transferring the vision foundation models to modalities other than natural images. While training foundation models and finetuning them on downstream tasks has been extensively studied~\cite{ye2024superanimal,pai2024foundation,cai2024pretrainable}, the potential of generalizing foundation models to novel image modalities is not fully explored. Arguably, transferring the foundation models to various input modalities like task transfer learning 
has the potential to unleash the power of the foundation model on specific sensors: we can utilize the advantages of sensors 
in capturing specific physical properties of objects in the world with a strong foundation model. 

The challenges for transferring vision foundation models to other image modalities come from two sides: the modality misalignment and the finetuning cost. A key challenge of cross-modality transfer learning comes from the modality gap: the captured physical signals and the data representation can be highly different, such as the dimensions, the dynamic ranges, and semantic information. Among many differences, the dimension misalignment is one of the major challenges, preventing people from finetuning on new modalities directly. A simple example is that RGB images capture the visible color of objects with three channels. In contrast, a polarization sensor can capture the polarization state of light with more than three channels, preventing it from utilizing the pretrained weights directly. The second challenge comes from the finetuning cost, which is increasing rapidly along with the quick growth of the model size of foundation models. To this end, a systematical analysis for applying different parameter-efficient finetuning strategies to cross-sensor transfer learning can be beneficial.

Researchers have attempted to explore cross-modal transfer learning in different modalities but most works focus on transferring a pretrained modality to another specific modality, including from language to vision~\cite{lu2022frozen,dinh2022lift} or protein sequences~\cite{vinod2023reprogramming}, from natural images to medical imaging~\cite{wu2023medical,ma2023medsam}. Few literatures have studied how to design a general cross-modal transfer framework for different image modalities. For example, Lu et al.~\cite{lu2022frozen} proposes a framework FPT for transferring pretrained transformers to different inputs. Most recently, Shen et al.~\cite{shen2023cross} propose a general cross-modal transfer learning framework for diverse modalities, including language, vision, tabular, $etc$. However, they do not carefully handle the modality misalignment~\cite{lu2022frozen} or require large computational cost~\cite{wu2023medical} or extra data~\cite{shen2023cross}. Besides, they do not take into account finetuning strategies, which is quite important in practice. In contrast, we study how to transfer the vision foundation model comprehensively, including handling modality misalignment and analyzing fine-tuning strategies.

To investigate this problem, we introduce SimMAT: a simple framework for modality-agnostic transfer learning from a vision foundation model to any imaging modality. SimMAT consists of a modality-agnostic layer and a pretrained vision foundation model. First, SimMAT is designed to accept any imaging modality as input for transfer learning. It does not require domain-specific knowledge, such as the relationship between the modality and natural images. With extensive exploratory experiments and analysis, we propose a simple and effective strategy to align the target modality and the pretrained vision modality in SimMAT. Secondly, we provide a comprehensive empirical study of parameter-efficient fine-tuning (PEFT) strategies on cross-modal transfer learning. Specifically, we compare the best performance of different strategies, including LoRA~\cite{hu2021lora}, MLP Adapter~\cite{houlsby2019parameter}, prompt tuning~\cite{zhu2023visual}, and full-finetuning. Results confirm that the performance of parameter-efficient finetuning could be better than full-finetuning when the training data is limited, which is consistent with the observations in in-modality finetuning.

In this paper, we focus on applying SimMAT to a recent vision foundation model Segment Anything Model (SAM)~\cite{kirillov2023segment} so that it can be used for segmentation in different image modalities. 
SAM is trained on 11 million images for a fundamental image segmentation task. To enable a fair comparison to study the transferring of SAM on novel modality, we build a dedicatedly designed segmentation benchmark that consists of datasets captured by various types of sensors, including polarization sensors, depth sensors, thermal sensors, and other types of sensors.

Extensive results demonstrate that SimMAT can achieve significant performance improvement across image modalities compared with models that are trained on specific modalities only. We find that SimMAT does improve the performance of other modalities despite these sensors capturing different physical properties in different representations. We hope our SimMAT can serve as a flexible and solid tool for transferring vision foundation models to other image modalities in different areas. Besides, we believe our findings can provide insights to explore the possibility of building a foundation model that processes any sensor modality for any task.

\newcommand{\exptrainingsctrach}{22.15}
\newcommand{\expsimmat}{53.88}

\newcommand{\mattrainingsctrach}{25.43}
\newcommand{\matrandomly}{58.89}
\newcommand{\matlinear}{63.96}
\newcommand{\mattranspose}{70.90}
\newcommand{\matours}{72.69}

\section*{Results}
\subsection*{Framework overview}
To explore the transferability of vision foundation models to other image modalities, we introduce \texttt{SimMAT}, a modality-agnostic transfer learning method, illustrated in Figure~\ref{fig:SAM-AIM}. SimMAT consists of a MAT layer $m$ and a foundation model $f$. SimMAT is designed to accept any type of image modality as input for transfer learning. It does not require domain-specific knowledge, such as the relationship between the modality and natural images. We select SAM as the representative vision foundation model $f$. The original module in SAM receives a three-dimensional RGB image to an embedding with $d$ dimensions. Given an input $\mathbf{x}$, the output $\mathbf{y}$ is obtained by:
\begin{align}
    \mathbf{y} = f(\mathbf{e}) = f(m(\mathbf{x})),
\end{align}
where $\mathbf{e}$ is the output embedding of our MAT layer. The MAT layer $m$ transfers a new input $\mathbf{x}$ with modality dimension $C$ to a modality embedding with the original vision embedding dimension $d$. In our experiments, we observe that different designs of MAT layers lead to significantly different model accuracy. For the foundation model $f$, we apply different finetuning strategies to study this problem, including parameter-efficient finetuning and full-finetuning.

\subsection*{Dataset construction}
\label{sec:AIM-Benchmark}
This section presents the details of our dataset. Since there is no existing benchmark that covers different types of modalities for the promotable segmentation task of SAM, we construct a new benchmark named Any Image Modality Segmentation (AIMS) benchmark. Specifically, we choose five representative sensors in different fields and their corresponding images as follows:
\begin{itemize}
    \item \textit{Polarization Images} capture the polarization state of the light. The polarization image is a nine-channel image. The polarization state is closely related to the shape and materials of objects and can be used for challenging tasks for conventional intensity cameras, such as camouflaged object detection, transparent object segment, reflection removal, $etc$.  We adopt RGBP-Glass~\cite{glass_dataset} and ZJU-RGBP~\cite{zjurgbp} in our benchmark. RGBP-Glass contains 3207 and 1304 images for training and evaluation, respectively. ZJU-RGBP includes 344 training images and 50 validation images.   
    \item \textit{Depth Images} capture scene geometry, which is commonly used in diverse applications, including robotics, autonomous driving, and computational photography. The depth image captured from the camera is a one-channel image. In our benchmark, we adopt the public NYUv2 dataset~\cite{Silberman:ECCV12}, which contains 1449 RGBD samples covering 40 categories. 
    
    \item \textit{HHA Images} are processed features obtained from depth images, which we analyze as a new modality~\cite{gupta2014learning}. The HHA encoding is a method for representing depth images in a way that captures additional geometric information beyond just depth. HHA uses three channels at each pixel to encode the horizontal disparity, the height above ground, and the angle with gravity.

    \item \textit{Thermal Images} capture thermal radiation coming from scenes or environments despite the weather and illumination conditions,
    which are commonly in various areas. The thermal images are usually one-channel. In our benchmark, we adopt the public Thermal-based glass segmentation dataset~\cite{huo2023glass}, which contains 5551 images with segmentation labels.
    
    \item \textit{NIR Images} can capture the light in near-infrared frequency, which are commonly used in low-light vision. The NIR (Near-Infrared) images are usually one-channel. We adopt the IVRG-NIR dataset~\cite{brown2011multi} in our benchmark, which consists of 477 NIR images and their ground truth.
\end{itemize}

We select these modalities as they capture significantly different properties of scenes compared with conventional intensity cameras, and they are quite different from each other. Besides, there are publicly available segmentation datasets for these modalities, and the effectiveness of the novel modality has been proven in previous works. Compared with the training data of RGB-based SAM, which contains 11 million images and more than 1 billion masks, most datasets have a limited number of training images and masks. The segmentation labels of SAM are instance-level segmentation. However, for some segmentation datasets, only semantic labels are provided, which is different from the requirement of the SAM training setting. Hence, post-processing is required to convert the ground truth format to the SAM training setting. Details are presented in the Supplementary Information.

\subsection*{Performance evaluation}
We evaluate SimMAT for segmentation transfer across modalities on our constructed dataset. Following the protocol of the interactive setting adopted in SAM~\cite{kirillov2023segment}, the center point of an instance is used as the default click prompt fed into the network. We adopt ViT-Base as the image encoder backbone of the pretrained SAM for all experiments. As the best learning rate can be different for each model, we sweep the learning rates and report the best performance for each model for a fair comparison. For all evaluated modalities, we only require the number of channels $C$ and then build our SimMAT for end-to-end training. Before exploring how to perform transfer learning, we first implement baseline approaches as references, $i.e.$, \textit{training from scratch}. The most naive baseline is to inherit the SAM architecture without pretrained weights and train the network only with the new modality data. While we understand it is quite challenging to train a Transformer model effectively with a small amount of data, we adopt this method as a baseline to keep experimental factors the same for reference. This baseline approach only achieves a low \exptrainingsctrach \% average mIoU on our benchmark.

\begin{figure}[t]
\centering
\includegraphics[width=1\textwidth]{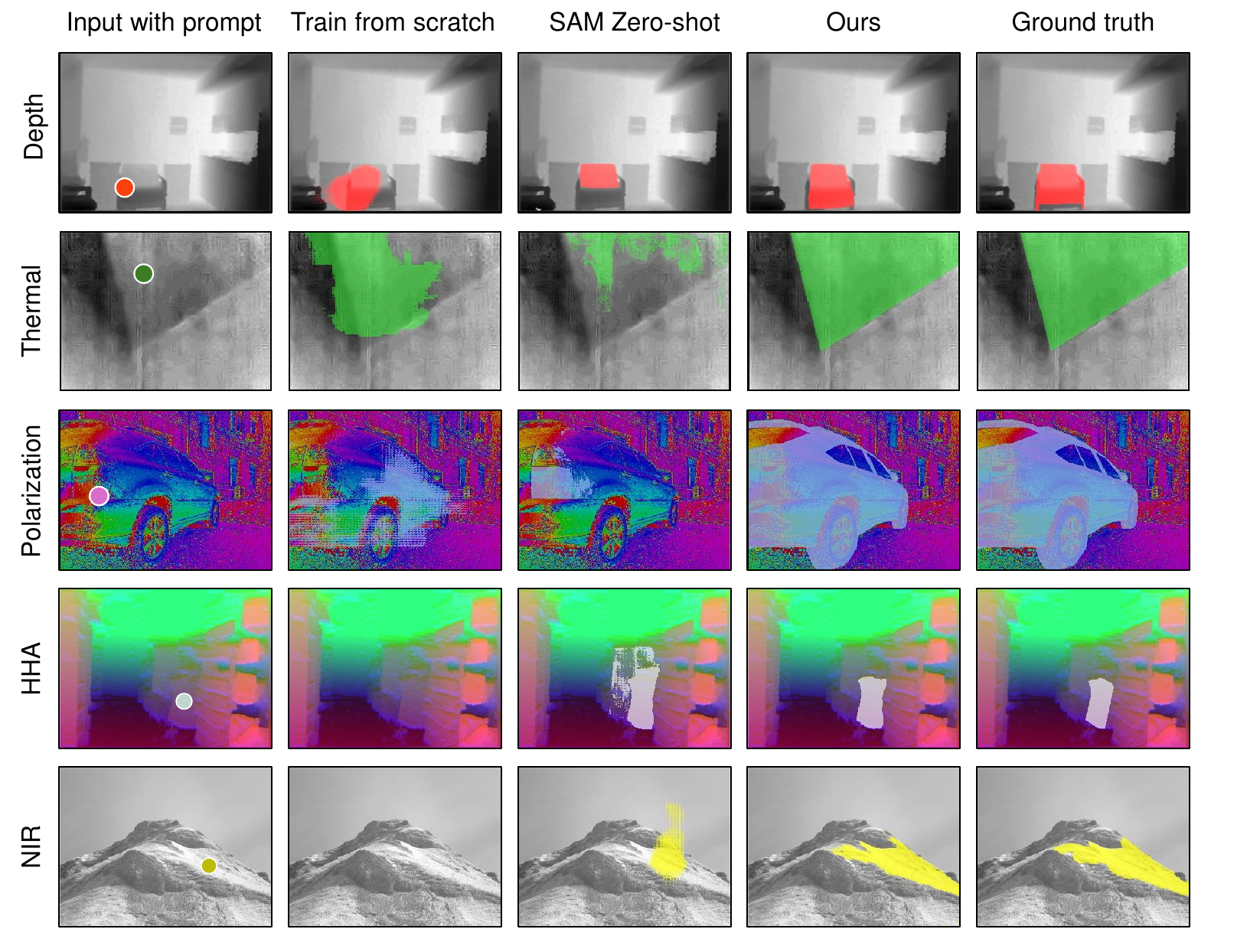}

\caption{\textbf{Qualitative Results.} We transfer the segment anything ability of SAM to different modalities, including segmentation from depth, thermal, polarization , HHA, and NIR images. The proposed method significantly improves segmentation quality compared to SAM zero-shot and training from scratch. }
\label{fig:qualitative_result}
\end{figure}

Training with SimMAT can achieve significantly better performance compared with training the models on specific data from scratch on our evaluated dataset. Figure~\ref{fig:baseline_cmp_num}(a) presents the results of our approach on different modalities. The results of training from scratch for all modalities are poor, which only gets \exptrainingsctrach\% mIoU on different modalities. As a comparison, training the model with SimMAT achieves \expsimmat\% mIoU, which is significantly better than training from scratch. This phenomenon is according to our expectations as the transformer is data-hungry and requires a large number of data for training. Since the dataset size for these sensors is usually small, they cannot train a good foundation model. This significant improvement demonstrates the potential and importance of cross-modal transfer learning from vision foundation models to other modalities in different fields. 
We further analyze the visual results to better understand the phenomenon in Figure~\ref{fig:qualitative_result}. The perceptual performance of training from scratch is poor, where the mask is inaccurate. As a comparison, training the model with SimMAT can obtain accurate and sharp segmentation results. We observe similar phenomena in all evaluated modalities, which demonstrates the effectiveness of our proposed SimMAT.

We further study a data representation that combines natural images and a paired new modality. This representation has been studied in previous methods as a multi-modality representation~\cite{zhu2023visual}.
However, in our setting, while we use both natural images and new modality, we \textit{assume} that we do not know the modality sequence: we do not adopt any RGB prior knowledge to process the RGB images individually so that it could be a special case of new modality named pseudo-new modality. Note that while this prior knowledge is easy to obtain, we just use these pseudo-new modalities to validate the effectiveness of our approach. Specifically, we shuffle the channels to avoid using domain knowledge. Since we have access to RGB images in this experiment, we provide an additional reference method of directly inputting the RGB image to SAM, which we denote as SAM zero-shot. As we can see in Figure~\ref{fig:baseline_cmp_num}(b), SAM zero-shot achieves reasonable performance but is far from our approach. As a comparison, our SimMAT framework with different finetuning strategies achieves much better performance compared with baselines. More qualitative results are presented in the supplementary materials due to limited space.

\subsection*{Controlled experiments of MAT}
\label{subsec:MATNet}
Dimension misalignment is an inevitable challenge in cross-modal transfer learning: the dimension of target modality can differ from the dimension of pretrained models, e.g., from vision to language~\cite{llava}, from natural images to medical images~\cite{wu2023medical}. While there are some commonly adopted naive strategies, such as using linear layers to change the dimensions, handling dimension misalignment with satisfying performance is still an open problem to solve up to date. In this section, we provide extensive ablation experiments here to analyze the design of our MAT layer. We just changed the design of the MAT layer for all experiments. Due to limited computing resources, all experiments are conducted using polarization images~\cite{glass_dataset} that consist of unpolarized intensity images, angle of linear polarization images, and degree of linear polarization images as the target new modality $\mathbf{x}$. Polarization images have nine channels. We load the pretrained weights for the foundation model and train the whole model jointly. We adopt LoRA~\cite{hu2021lora} to train these models.

Table~\ref{table:MAT_design} shows different existing methods for solving the dimension misalignments in different tasks. While some representative works like ORCA~\cite{shen2023cross} and BLIP-2~\cite{li2023blip} propose methods for aligning dimensions, they require source-target paired data (e.g., image-text pair), which are not available under our setting. Hence, their modules are not applicable. There are some direct dimension alignment strategies adopted in priors works. We apply these methods in our SimMAT framework but they cannot achieve satisfying performance. \textit{(1) Randomly initialized MAT layer.} We start by replacing the original patch embedding with a randomly initialized projection layer. This strategy is quite direct and naive, which has been commonly used in prior works~\cite{sun2019rtfnet,singh2023depth,lu2022frozen}. Compared to training from scratch, this implementation fully utilizes the pretrained vision model weights. As a result, mIoU is improved from \mattrainingsctrach \% to \matrandomly \% compared to training from scratch, validating the potential of modality-agnostic transfer learning. However, the performance is significantly worse than our proposed strategy achieving \matours  \% mIoU. \textit{(2) Inherited vision embedding with linear layer.} Another common strategy to align different dimensions is using a linear layer ($i.e.$, a fully connected layer). Many prior works adopt this strategy due to its simplicity and efficiency, including LLaVa~\cite{llava}. With this simple strategy, we improve the mIoU score from 58.89\% to 63.96\% compared with the randomly initialized MAT layer. However, the performance is still worse than our 72.69\%. We believe it is because the transformation of the linear layer is too simple to align two different image modalities, preventing it from achieving satisfying performance. \textit{(3) Transpose to batch dimension.} Another strategy is to transpose the feature dimension to the batch dimension and process them separately, such as the method used in MedicalSAM~\cite{wu2023medical}. Interestingly, we observe this implementation can achieve better performance than the prior two versions. Specifically, it gets 70.90\% on our evaluated dataset, which is close to our results. However, it suffers from a practical resource problem: it is inefficient for some image modalities. Specifically, for single-channel images, the FLOPs of this method are similar to ours. However, it uses around 9$\times$ FLOPs compared with our MAT layer when using polarization images, which makes it quite challenging to train the models for many areas.

We present a simple yet effective MAT layer for aligning new modalities. As shown in Table~\ref{table:MAT_design}, our design achieves 72.69\% mIoU, which is the best among different strategies. In addition, our MAT is also quite efficient. Different from these methods of building a complex module to bridge the modality gap, we find the two key factors to utilize the vision embedding layer for a new modality. First, the mapping from the novel modality to the pretrained RGB feature space is non-linear. Introducing a linear projection layer fails to achieve satisfactory performance due to limited mapping ability. Instead, SimMAT stacks convolutional layers with ReLU as an intermediate non-linear activation function. A similar conclusion is observed in contrastive learning~\cite{chen2020simple,moco}: replacing the linear layer with MLP as the projection head can achieve better performance. Second, the receptive field is important for novel imaging modalities. It helps capture the cues from neighbor regions and benefit the pixel to learn more rich context; such observation has been well studied and verified in the RGB modality, which inspired us to enlarge the receptive field of SimMAT by setting the convolutional kernel size. Specifically, we stack \textit{n} convolutional layers with \textit{k} kernel size and dimension \textit{d}, we set \{\textit{n}, \textit{k}, \textit{d}\} to \{2, 3, 64\} in default since it achieves the best performance, more detailed experiment results are present in supplemental materials.

\subsection*{Empirical analysis of finetuning strategies}
\label{sec:finetuning}

In this part, we conduct empirical experiments to explore which strategy works better for modality-agnostic transfer learning. After handling the dimension misalignment with our designed MAT layer, SimMAT can adopt existing finetuning strategies easily like in-modality transfer learning. However, it is not validated systematically on many sensors in different areas, such as polarization~\cite{lei2020polarized}. We mainly explore two commonly used finetuning styles here: (1) Full Finetuning (FFT): full finetuning is commonly used as it usually achieves satisfying performance easily. Following this setting, we make both the MAT layer and pretrained backbone learnable. (2) Parameter-efficient Finetuning (PEFT): as the parameters of foundation models are usually very large, full finetuning a model might be extremely resource-hungry. PEFT strategies usually fix the original parameters and introduce a small amount of learnable new parameters. In our experiments, we select representative parameter-efficient finetuning methods, including LoRA,  MLP Adapter, prompt tuning, $etc$.

\begin{figure*}[t!]
  \centering
  \includegraphics[width=0.55\linewidth]{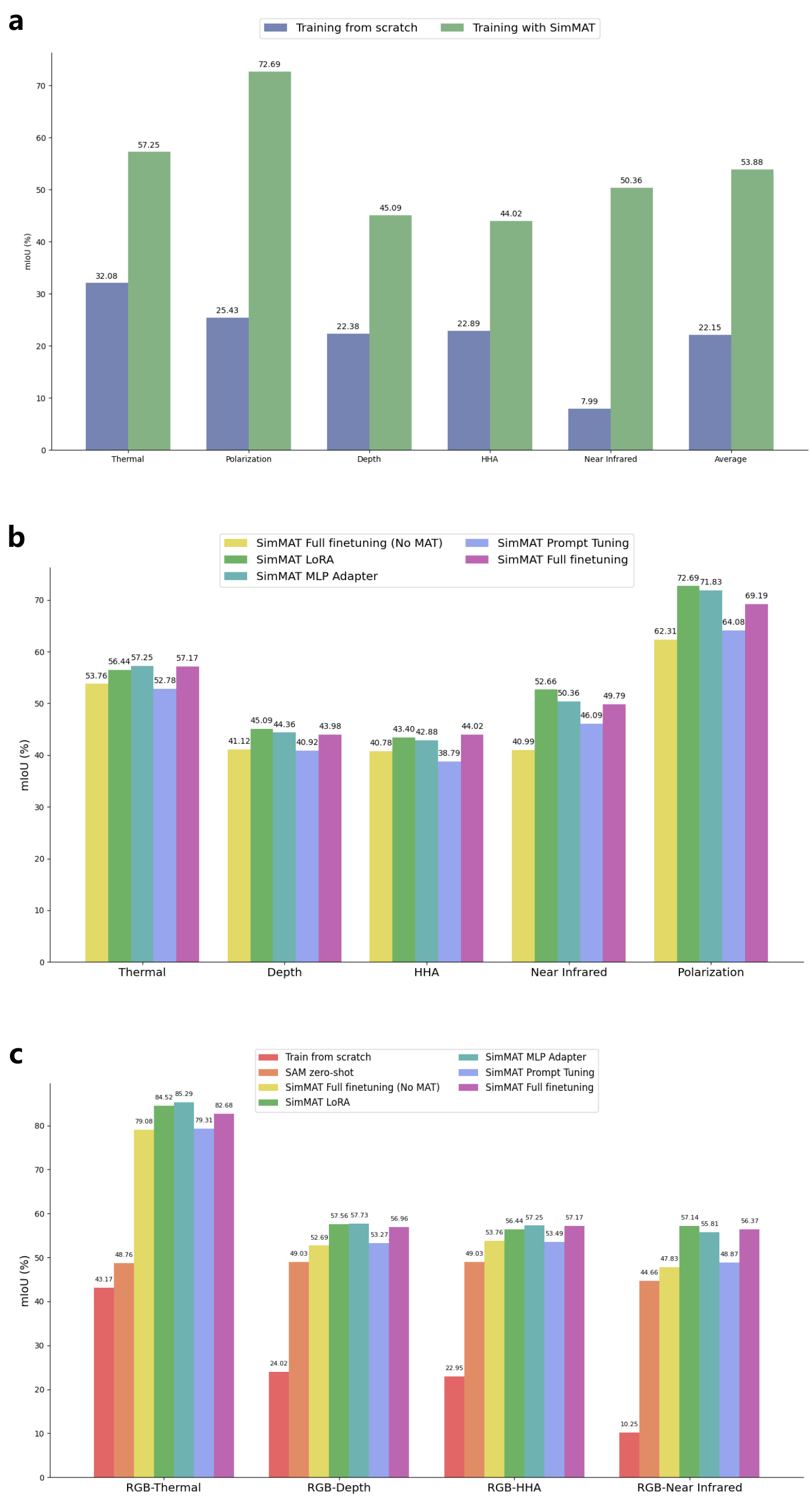}
    \caption{\textbf{Performance Evaluation on Different Modalities.} {\textbf{a}.} The proposed method SimMAT improves the segmentation performance significantly on all evaluated modalities compared with training the models from scratch. Specifically, SimMAT improves the mIoU from 22.15\% to 53.88\% for all evaluated modalities on average.  Besides, the peak performance between finetuning and parameter-efficient finetuning is similar. 
    \textbf{b}. Results on Pseudo New Modalities. We combine natural images with a novel image modality as a pseudo new modality: note that we do not use the information that which three channels are for natural images and which channels are for new modalities.  For example, our MAT is effective in improving the finetuning performance on all evaluated pseudo new modalities. Besides, the peak performance between finetuning and parameter-efficient finetuning is similar.
    \textbf{c}. We provide controlled experiments for different finetuning strategies on new modalities. Parameter-efficient finetuning strategies can achieve comparable performance compared with full finetuning by using much less trainable parameters.
  }
  \label{fig:baseline_cmp_num}
\end{figure*}

Figure~\ref{fig:baseline_cmp_num}(b) presents the results of different finetuning strategies. We sweep the learning rate for each finetuning method and choose the best result for comparison. All finetuning strategies can improve the segmentation performance compared with training from scratch. LoRA and Adapter can achieve similar performance with full finetuning while they only use much fewer trainable parameters. Besides, while prompt tuning can improve the performance compared with training the models from scratch, the results fall below that of the other two PEFT methods despite having a close number of trainable parameters, as shown in Figure~\ref{fig:baseline_cmp_num}(b). We believe this is attributed to the initial noise brought by the prompts embedding. It fails to find an initialization that ensures prompt embeddings do not disturb the model output at the first forward pass. In comparison, the effects of both LoRA and MLP adapter on the model can be initialized as zero.

\begin{figure*}[t!]
  \centering
  \includegraphics[width=1\linewidth]{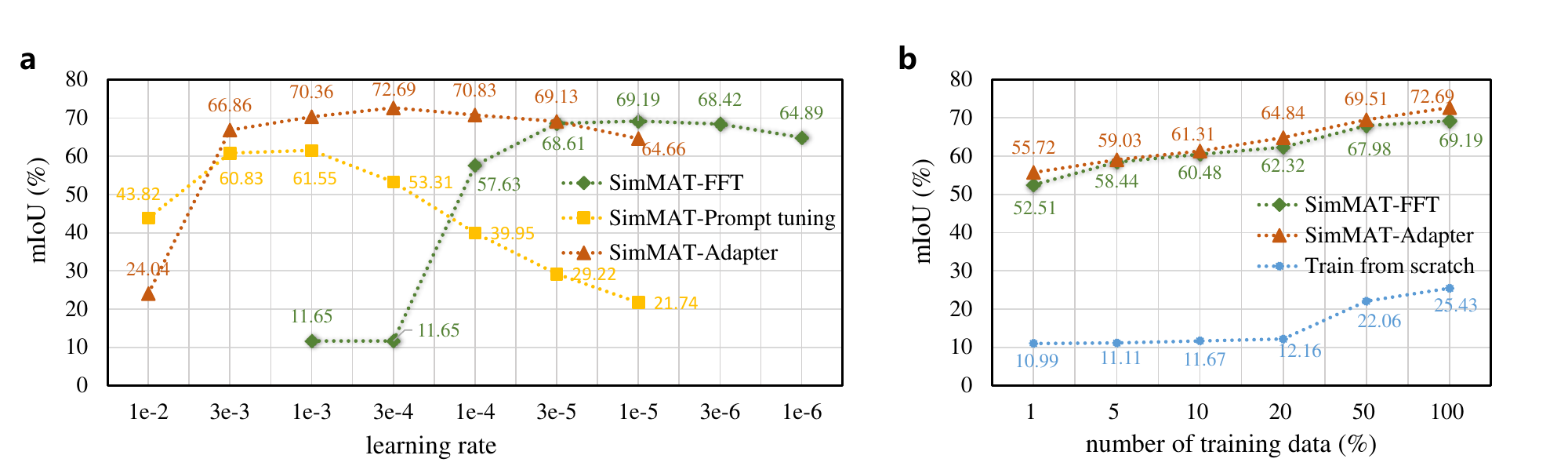}
    \caption{\textbf{The Effect of Learning Rate and Training Data Size.} The models are evaluated on the polarization modality. \textbf{a}. the full fine-tuning and parameter efficient tuning achieve peak performance in different learning rates. \textbf{b}. increasing the scale of training data brings consistent performance improvement across different training strategies.}
  \label{fig:lr_data_effect}
    \end{figure*}

We further study the effect of learning rate for different finetuning strategies. Prior works~\cite{hu2021lora,jia2022visual} noticed that different tuning strategy holds different best learning rates, we observe consistent results as presented in Figure~\ref{fig:lr_data_effect}(a). 
Full finetuning achieves a peak performance 69.19\% mIoU at lr=1e-5, while parameter-efficient finetuning achieves the best 72.69\% mIoU at lr=3e-4. 
We suspect the reason is the number of trainable parameters. Full finetuning makes all parameters learnable; a small learning rate prevents the model from deviating far away from the pretrained weights. 
While LoRA or MLP Adapter with only 4\% trainable parameters demands a larger learning rate for efficient learning. 

To study the relationship between the number of finetuned images and the pretrained model, we provide a controlled experiment. We split the training set randomly according to different ratios. Figure~\ref{fig:lr_data_effect}(b) shows the results. We notice that using RGB-based pretrained SAM can significantly improve the performance on different image modalities, especially when the training images of specific modalities are limited.

\section*{Discussion}
Foundation models (Large Models) have revolutionized artificial intelligence areas, such as ChatGPT~\cite{ouyang2022training} in Natural Language Processing and SAM (Segment Anything Model) in Computer Vision. Driven by the availability of large-scale image data, several foundation models have recently been proposed~\cite{clip,controlnet,kirillov2023segment,lisa} for vision tasks including image recognition, segmentation, conditional generation, etc. As a result, numerous downstream tasks can achieve impressive performance. 
Nevertheless, except for conventional cameras, the available data of many image sensors is not large enough, preventing applications in different areas from benefitting from the significant progress of foundation models. Transferring the ability of vision foundation models to new data-limited image modalities is promising, but this line of work has not been fully explored or studied. 

In this work, we confirm the potential of modality-agnostic cross-modal transfer learning from vision foundation models to other image modalities beyond natural images. To this end, we introduce a training paradigm SimMAT to study this problem. We conduct extensive exploratory experiments to propose a good MAT layer for receiving different types of new modalities. We explore different finetuning strategies and report our observations. Based on these experiments, we validate the transfer performance of our proposed SimMAT through a vision foundation model SAM on a variety of sensors. The significant margins achieved by SimMAT suggest that the generic cross-modal transfer learning is still underexplored. We envision SimMAT to be useful for other vision foundation models and other unevaluated modalities that are not studied in this work.

While achieving substantial improvements, we believe the upper bound of modality-agnostic transfer learning is not achieved with our method, implying rich future research in this direction. Possible research directions are described as follows: \textbf{(1) Domain-specific knowledge.} We argue using domain-specific knowledge is always a good choice for improving performance,  but it does not conflict with our modality-agnostic SimMAT. Designing domain-specific strategies can cost more time and effort at the beginning. In contrast, SimMAT can be applied to validate the effectiveness of a novel sensor for specific sensors much more efficiently. With the positive validation from SimMAT, researchers can then focus on combining domain-specific knowledge. {\textbf{(2) Why not collect more data?}} Collecting more data is one of the best ways to train a stronger foundation model~\cite{rombach2021highresolution, kirillov2023segment}. We believe this argument works for most image modalities as well. Alternative methods include creating more synthetic data for training, but it is non-trivial to synthesize other modalities. However, even if data would be free to collect, we argue our proposed SimMAT can be used to validate the effectiveness of a sensor efficiently with a low requirement for the training data and training cost. 
{\textbf{(3) Zero-shot modality transfer?}} While many existing foundation models demonstrate impressive {zero-shot} performance on new tasks, it is still extremely challenging for them to achieve satisfying zero-shot performance on many new modalities. We believe training a MAT layer is useful and necessary at this stage since the features are quite different between natural images and other sensors.

\section*{Method}

In this section, we present the setting of our methods, including the details of the data, the foundation models, and the training protocols.  
\subsection*{Related work} 
Unimodal transfer learning is commonly used in computer vision. Transfer learning first pretrains the model to learn prior knowledge and then fine-tunes it on another downstream task. It has shown to be effective in various areas~\cite{zamir2018taskonomy}. Most transfer learning is conducted on the same input modality. For example, the models are first trained on the ImageNet~\cite{deng2009imagenet} in a contrastive learning~\cite{chen2020simple, moco, moco-v2, moco-v3, BYOL} or a masking inpainting way~\cite{he2022masked, simmim, beit} and then used for the downstream task with RGB image input. Besides, the pertaining model works well in other scenarios like predictions of RNA secondary structure~\cite{singh2019rna}, metal-organic framework~\cite{kang2023multi, pan2023transfer} and fault slip~\cite{wang2021predicting}.

However, the modalities suffering from limited training data fail to perform pretrain-tuning paradigm. For example, modalities like polarization, structured light~\cite{choi2023neural}, and event camera~\cite{gehrig2024low,guo2024eventlfm}, which have proven instrumental in 3D imaging and auto-driving, but there is no large scale data for pertaining. Cross-modal transfer learning is a potential way to solve this problem, which has been studied, but most research explores this problem in a modality-specific style for a specific pair of modalities. For example, Radhakrishnan et al.\cite{radhakrishnan2023transfer} study transfer learning on image classification and virtual drug screening applications. Many works~\cite{lu2022frozen,dinh2022lift,llava,li2024llava} study the transferability from language models to vision. Zhang et al. employ the vision-language foundation model for biomedical tasks~\cite{zhang2024generalist}.
Vinod et al. ~\cite{vinod2023reprogramming} attempts to apply language models to protein sequences~\cite{vinod2023reprogramming}. Wu et al.~\cite{wu2023medical} and Ma et al.~\cite{ma2023medsam} attempt to transfer the vision segmentation model to medical imaging. 
Domain adaptation aims to transfer source domain knowledge to the target domain. Numerous methods are proposed to reduce the cross-domain discrepancy through adversarial learning~\cite{da_adverse_learning_1,da_adverse_learning_2,da_adverse_learning_3}, or introduce pseudo labels with self-supervised  learning~\cite{da_self_learning_1,da_self_learning_2,da_self_learning_3}.
Different from these methods considering the domain gap in RGB images, we try to alleviate the modality gap between RGB images and other image modalities, it is more challenging since the physical forward model changes between the image channels.
While heterogeneous domain adaptation~\cite{hda_1,hda_2} extensively discusses the feature space change, they usually assume the source training data is available.

Few works~\cite{lu2022frozen} study a general and modality-agnostic cross-modal transferring workflow of transferring the knowledge from pretrained data to downstream tasks. Lu et al.~\cite{lu2022frozen} proposes a framework FPT for transferring pretrained transformers to different inputs. However, they only study pretrained language models in a frozen way. Most recently, Shen et al.~\cite{shen2023cross} propose a general cross-modal transfer learning framework for diverse modalities, including language, vision, tabular, $etc$. Nevertheless, they do not take into account finetuning strategies, which is quite important in practice. 
In this paper, we are interested in exploring the transferability of the vision foundation model by investigating modality misalignment and finetuning strategies comprehensively.

Parameter-efficient finetuning is also closely related to our research. Fully finetuning a large transformer model costs large GPU memory and training time. Parameter-efficient finetuning solves this problem by freezing the pretrained foundation model and only finetuning a small number of parameters, which has been shown to achieve comparable or even better performance than fully finetuning. It was first proposed in the natural language processing task~\cite{houlsby2019parameter,hu2021lora} and then explored in the computer vision task~\cite{chen2022adaptformer}. Visual prompt tuning~\cite{jia2022visual} adds some learnable tokens before each transformer block. Visual adapter~\cite{chen2022adaptformer} inserts small multilayer perceptrons (MLPs) to the feed-forward network in a residual way~\cite{he2016deep}. Prefix-tuning~\cite{li2021prefix} adds a few parameters before each multi-head attention layer. LoRA~\cite{hu2021lora} optimize the rank-decomposition matrices of layers' change and achieve zero inference latency. 

Unified architectures and learning algorithms for various data modalities have emerged recently. Designing a foundation model~\cite{bommasani2021opportunities, internimage, llava, imagebind} for various modalities becomes a goal for the community. The transformer architecture~\cite{vaswani2017attention} has been proven to be very effective in different domains, including natural language~\cite{kenton2019bert,brown2020language,touvron2023llama}, vision~\cite{dosovitskiy2020image,liu2021swin,wang2021not}, point clouds~\cite{guo2021pct,zhao2021point,wu2022point}, audio~\cite{gong2021audio,chen2022hts,verma2021audio}, and so on. Perceiver~\cite{jaegle2021perceiver} is proposed for the general perception of various types of data modalities. PerceiverIO~\cite{jaegle2021perceiverio} unify the input and output structures, which demonstrate the effectiveness of images and videos. Tamkin et al.~\cite{tamkin2021dabs} construct a benchmark for domain-agnostic self-supervised learning algorithms~\cite{wu2023randomized,wang2023internimage}, including natural images, language, and sensors. Meta-transformer~\cite{zhang2023meta} uses a unified encoder for 12 modalities with each modality using a specific encoding way. Our SimMAT is also designed to handle different image modalities in a modality-agnostic formulation.

\subsection*{Framework architecture of SimMAT} 
\begin{figure}[t]
\centering
\includegraphics[width=1\textwidth]{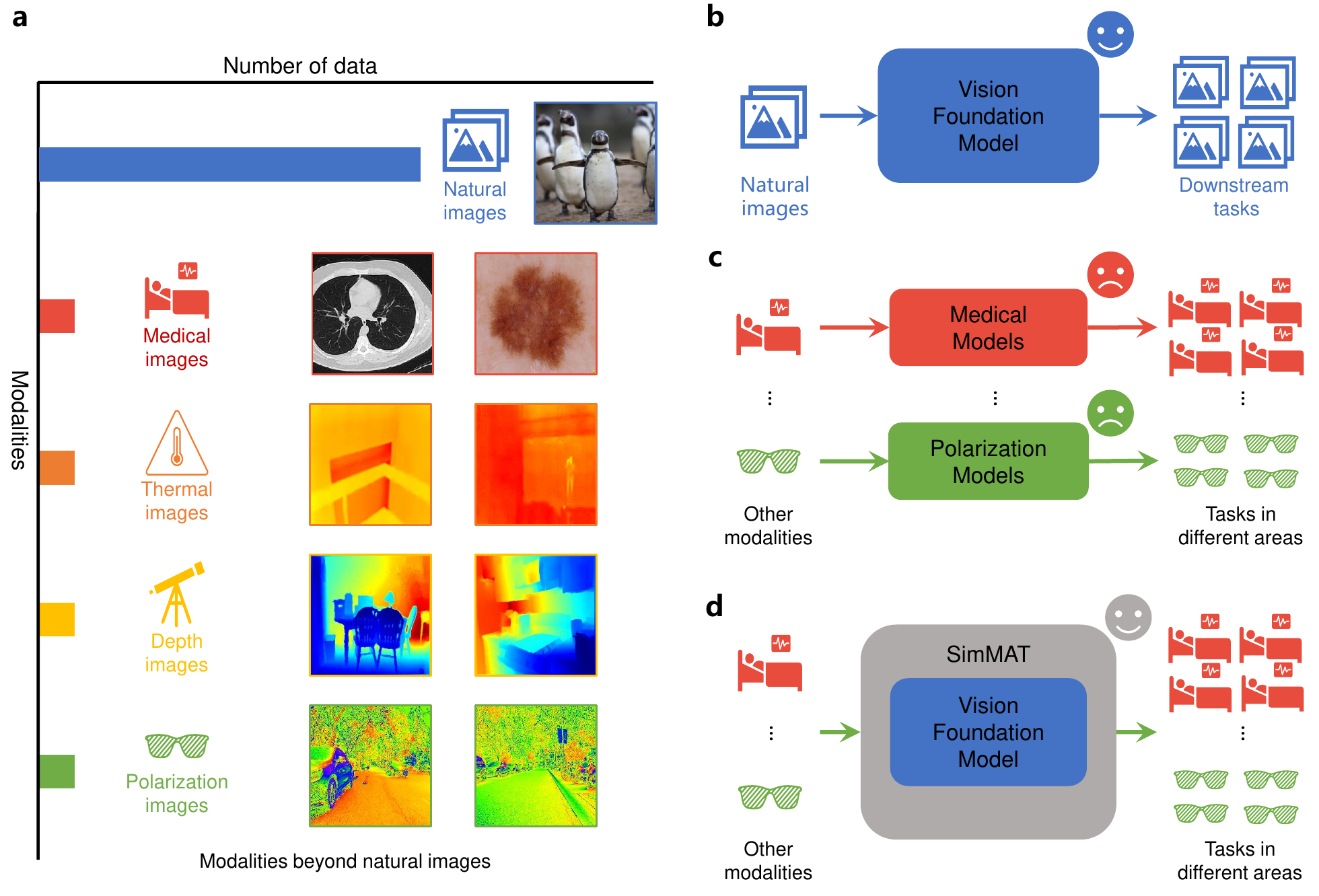}

\caption{\textbf{Transferability Across Modalities.} \textbf{a}, the number of natural images is significantly larger than images in other modalities in different areas, including medical imaging, thermal images, depth images, and polarization images. \textbf{b}, natural images can train vision foundation models, which can be applied to achieve strong performance on different downstream tasks. \textbf{c}, it is very challenging for other modalities to benefit from training foundation models due to limited data. \textbf{d}, our proposed SimMAT explores the transferability from the pretrained vision foundation model to different modalities. }
\label{fig:framework}
\end{figure}

\begin{figure}
\centering
\includegraphics[width=1\textwidth]{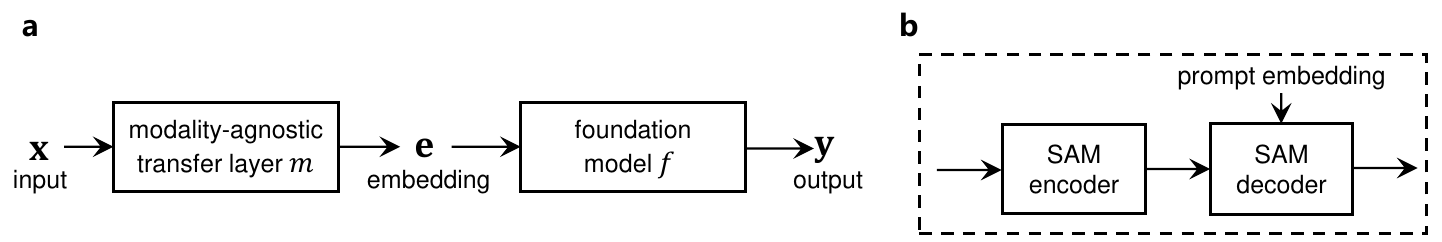}    
\caption{\textbf{Details of SimMAT.} \textbf{a}. SimMAT receives new modality  $\mathbf{x}$ as input and pass it through a modality-agnostic transfer layer $m$ to obtain an embedding $\mathbf{e}$. The embedding matches the dimension of a pretrained foundation model $f$, and then we obtain the output $\mathbf{y}$. The input and foundation are designed in a generic formulation for different modalities and foundation models. \textbf{b}. in this work, we select SAM as a representative foundation model for a detailed study. }
\label{fig:SAM-AIM}
\end{figure}

We propose a simple yet effective modality-agnostic transfer learning framework to transfer the ability of vision foundation models to sensors in different applications, which we name SimMAT. As shown in Fig.~\ref{fig:framework}, our framework is inspired by the attractive finetuning performance of in-modality transfer learning, from a pretrained task to different downstream tasks. Fig.~\ref{fig:SAM-AIM} shows the framework of SimMAT, which consists of a MAT layer and a foundation model. Different from in-modality transfer learning, novel image modalities captured from alternative sensors may have different dimensions, which prevents us from using existing finetuning methods directly. To address this problem, \texttt{SimMAT} introduces a modality-agnostic transfer layer (MAT). To keep our framework simple, the only difference is the MAT layer compared with in-modality finetuning. Considering a new input data $\mathbf{x} \in \mathbb{R}^{H\times W\times C}$ that is captured from a novel sensor, the dimension $C$ is a modality-specific value (e.g., $C=3$ for a RGB image and $C=1$ for a depth image). The design of MAT is open. In this paper, we propose a compact design of {MAT} layer. This module $m$ requires a pretrained RGB embedding layer $m_v$ and the novel modality dimension $C$ as input. As for the backbone, we load the pretrained weights from the vision foundation model directly. We will add a few trainable parameters if we use the parameter-efficient finetuning strategies, similar to in-modality finetuning.

\subsection*{Experimental setup} 
Visual foundation models have developed very fast~\cite{xdecoder,4m,seem,wang2023internimage,rombach2021highresolution}. This paper selects Segment Anything Model (SAM)~\cite{kirillov2023segment} as a backbone for exploring experiments as it is one of the most representative foundation models in computer vision. SAM has three components: an image encoder, a prompt encoder, and a mask decoder. The image encoder receives image patches as input and computes image features. The prompt encoder embeds prompts, $i.e.$, points, boxes, text, or masks. Both image features and prompt embedding are fed into a lightweight mask decoder to obtain mask predictions.
The released SAM model is trained on the large-scale SA-1B dataset, which contains over 1 billion
automatically generated masks (400× more masks than any existing segmentation datasets)
and 11 million images. Several works~\cite{ji2023sam, tang2023camouflage, chen2023samfail, ma2023medsam} focus on adapting the SAM to different domains of RGB images, while we use SAM as the vision foundation model to explore the modality transfer task. Although some works~\cite{wu2023medical, cen2023sad} have discussed the SAM adaption with specific modality ($e.g.$, MRI, depth), we are toward a more general setting handling an arbitrary modality.

We train the model using the same loss function of the vision foundation model ($i.e.$, SAM in this paper). In our experiments, we investigate different finetuning strategies, including LoRA~\cite{hu2021lora}, MLP Adapter~\cite{chen2022adaptformer}, full finetuning, $etc$. We train the model on each modality respectively. We implement the SimMAT using PyTorch. More details for the training can be found in the supplementary materials.

\section*{Data Availability}
The source images used throughout this work are publicly available. All captured data used to generate the findings in this work will be made public.

\section*{Code Availability}
Our code will be publicly available at \href{https://github.com/mt-cly/SimMAT/}{https://github.com/mt-cly/SimMAT/}.

\bibliography{reference}

\begin{thebibliography}{100}
\urlstyle{rm}
\expandafter\ifx\csname url\endcsname\relax
  \def\url#1{\texttt{#1}}\fi
\expandafter\ifx\csname urlprefix\endcsname\relax\def\urlprefix{URL }\fi
\expandafter\ifx\csname doiprefix\endcsname\relax\def\doiprefix{DOI: }\fi
\providecommand{\bibinfo}[2]{#2}
\providecommand{\eprint}[2][]{\url{#2}}

\bibitem{kirillov2023segment}
\bibinfo{author}{Kirillov, A.} \emph{et~al.}
\newblock \bibinfo{title}{Segment anything}.
\newblock In \emph{\bibinfo{booktitle}{Proceedings of the IEEE/CVF International Conference on Computer Vision}}, \bibinfo{pages}{4015--4026} (\bibinfo{year}{2023}).

\bibitem{bai2024sequential}
\bibinfo{author}{Bai, Y.} \emph{et~al.}
\newblock \bibinfo{title}{Sequential modeling enables scalable learning for large vision models}.
\newblock In \emph{\bibinfo{booktitle}{Proceedings of the IEEE/CVF Conference on Computer Vision and Pattern Recognition}}, \bibinfo{pages}{22861--22872} (\bibinfo{year}{2024}).

\bibitem{bommasani2021opportunities}
\bibinfo{author}{Bommasani, R.} \emph{et~al.}
\newblock \bibinfo{journal}{\bibinfo{title}{On the opportunities and risks of foundation models}}.
\newblock {\emph{\JournalTitle{arXiv preprint arXiv:2108.07258}}}  (\bibinfo{year}{2021}).

\bibitem{kenton2019bert}
\bibinfo{author}{Kenton, J. D. M.-W.~C.} \& \bibinfo{author}{Toutanova, L.~K.}
\newblock \bibinfo{title}{Bert: Pre-training of deep bidirectional transformers for language understanding}.
\newblock In \emph{\bibinfo{booktitle}{Proceedings of naacL-HLT}}, vol.~\bibinfo{volume}{1}, \bibinfo{pages}{2} (\bibinfo{year}{2019}).

\bibitem{brown2020language}
\bibinfo{author}{Brown, T.} \emph{et~al.}
\newblock \bibinfo{journal}{\bibinfo{title}{Language models are few-shot learners}}.
\newblock {\emph{\JournalTitle{Advances in neural information processing systems}}} \textbf{\bibinfo{volume}{33}}, \bibinfo{pages}{1877--1901} (\bibinfo{year}{2020}).

\bibitem{ma2024segment}
\bibinfo{author}{Ma, J.} \emph{et~al.}
\newblock \bibinfo{journal}{\bibinfo{title}{Segment anything in medical images}}.
\newblock {\emph{\JournalTitle{Nature Communications}}} \textbf{\bibinfo{volume}{15}}, \bibinfo{pages}{654} (\bibinfo{year}{2024}).

\bibitem{zhou2024pre}
\bibinfo{author}{Zhou, J.} \emph{et~al.}
\newblock \bibinfo{journal}{\bibinfo{title}{Pre-trained multimodal large language model enhances dermatological diagnosis using skingpt-4}}.
\newblock {\emph{\JournalTitle{Nature Communications}}} \textbf{\bibinfo{volume}{15}}, \bibinfo{pages}{5649} (\bibinfo{year}{2024}).

\bibitem{yang2024vision}
\bibinfo{author}{Yang, Z.} \emph{et~al.}
\newblock \bibinfo{journal}{\bibinfo{title}{A vision chip with complementary pathways for open-world sensing}}.
\newblock {\emph{\JournalTitle{Nature}}} \textbf{\bibinfo{volume}{629}}, \bibinfo{pages}{1027--1033} (\bibinfo{year}{2024}).

\bibitem{gehrig2024low}
\bibinfo{author}{Gehrig, D.} \& \bibinfo{author}{Scaramuzza, D.}
\newblock \bibinfo{journal}{\bibinfo{title}{Low-latency automotive vision with event cameras}}.
\newblock {\emph{\JournalTitle{Nature}}} \textbf{\bibinfo{volume}{629}}, \bibinfo{pages}{1034--1040} (\bibinfo{year}{2024}).

\bibitem{yako2023video}
\bibinfo{author}{Yako, M.} \emph{et~al.}
\newblock \bibinfo{journal}{\bibinfo{title}{Video-rate hyperspectral camera based on a cmos-compatible random array of fabry--p{\'e}rot filters}}.
\newblock {\emph{\JournalTitle{Nature Photonics}}} \textbf{\bibinfo{volume}{17}}, \bibinfo{pages}{218--223} (\bibinfo{year}{2023}).

\bibitem{merchant2023scaling}
\bibinfo{author}{Merchant, A.} \emph{et~al.}
\newblock \bibinfo{journal}{\bibinfo{title}{Scaling deep learning for materials discovery}}.
\newblock {\emph{\JournalTitle{Nature}}} \textbf{\bibinfo{volume}{624}}, \bibinfo{pages}{80--85} (\bibinfo{year}{2023}).

\bibitem{pan2009survey}
\bibinfo{author}{Pan, S.~J.} \& \bibinfo{author}{Yang, Q.}
\newblock \bibinfo{journal}{\bibinfo{title}{A survey on transfer learning}}.
\newblock {\emph{\JournalTitle{IEEE Transactions on knowledge and data engineering}}} \textbf{\bibinfo{volume}{22}}, \bibinfo{pages}{1345--1359} (\bibinfo{year}{2009}).

\bibitem{wu2023medical}
\bibinfo{author}{Wu, J.} \emph{et~al.}
\newblock \bibinfo{journal}{\bibinfo{title}{Medical sam adapter: Adapting segment anything model for medical image segmentation}}.
\newblock {\emph{\JournalTitle{arXiv preprint arXiv:2304.12620}}}  (\bibinfo{year}{2023}).

\bibitem{huang2023polarization}
\bibinfo{author}{Huang, X.} \emph{et~al.}
\newblock \bibinfo{journal}{\bibinfo{title}{Polarization structured light 3d depth image sensor for scenes with reflective surfaces}}.
\newblock {\emph{\JournalTitle{Nature Communications}}} \textbf{\bibinfo{volume}{14}}, \bibinfo{pages}{6855} (\bibinfo{year}{2023}).

\bibitem{lei2022shape}
\bibinfo{author}{Lei, C.} \emph{et~al.}
\newblock \bibinfo{title}{Shape from polarization for complex scenes in the wild}.
\newblock In \emph{\bibinfo{booktitle}{Proceedings of the ieee/cvf conference on computer vision and pattern recognition}}, \bibinfo{pages}{12632--12641} (\bibinfo{year}{2022}).

\bibitem{lei2020polarized}
\bibinfo{author}{Lei, C.} \emph{et~al.}
\newblock \bibinfo{title}{Polarized reflection removal with perfect alignment in the wild}.
\newblock In \emph{\bibinfo{booktitle}{Proceedings of the IEEE/CVF conference on computer vision and pattern recognition}}, \bibinfo{pages}{1750--1758} (\bibinfo{year}{2020}).

\bibitem{sun2022seeing}
\bibinfo{author}{Sun, Z.}, \bibinfo{author}{Wang, J.}, \bibinfo{author}{Wu, Y.} \& \bibinfo{author}{Nayar, S.}
\newblock \bibinfo{title}{Seeing far in the dark with patterned flash}.
\newblock In \emph{\bibinfo{booktitle}{European Conference on Computer Vision}}, \bibinfo{pages}{709--727} (\bibinfo{organization}{Springer}, \bibinfo{year}{2022}).

\bibitem{gallego2020event}
\bibinfo{author}{Gallego, G.} \emph{et~al.}
\newblock \bibinfo{journal}{\bibinfo{title}{Event-based vision: A survey}}.
\newblock {\emph{\JournalTitle{IEEE transactions on pattern analysis and machine intelligence}}} \textbf{\bibinfo{volume}{44}}, \bibinfo{pages}{154--180} (\bibinfo{year}{2020}).

\bibitem{dong2016hyperspectral}
\bibinfo{author}{Dong, W.} \emph{et~al.}
\newblock \bibinfo{journal}{\bibinfo{title}{Hyperspectral image super-resolution via non-negative structured sparse representation}}.
\newblock {\emph{\JournalTitle{IEEE Transactions on Image Processing}}} \textbf{\bibinfo{volume}{25}}, \bibinfo{pages}{2337--2352} (\bibinfo{year}{2016}).

\bibitem{tseng2021neural}
\bibinfo{author}{Tseng, E.} \emph{et~al.}
\newblock \bibinfo{journal}{\bibinfo{title}{Neural nano-optics for high-quality thin lens imaging}}.
\newblock {\emph{\JournalTitle{Nature communications}}} \textbf{\bibinfo{volume}{12}}, \bibinfo{pages}{6493} (\bibinfo{year}{2021}).

\bibitem{wang2024sub}
\bibinfo{author}{Wang, R.} \emph{et~al.}
\newblock \bibinfo{journal}{\bibinfo{title}{Sub-surface thermal measurement in additive manufacturing via machine learning-enabled high-resolution fiber optic sensing}}.
\newblock {\emph{\JournalTitle{Nature Communications}}} \textbf{\bibinfo{volume}{15}}, \bibinfo{pages}{7568} (\bibinfo{year}{2024}).

\bibitem{mao2024multimodal}
\bibinfo{author}{Mao, Q.}, \bibinfo{author}{Liao, Z.}, \bibinfo{author}{Yuan, J.} \& \bibinfo{author}{Zhu, R.}
\newblock \bibinfo{journal}{\bibinfo{title}{Multimodal tactile sensing fused with vision for dexterous robotic housekeeping}}.
\newblock {\emph{\JournalTitle{Nature Communications}}} \textbf{\bibinfo{volume}{15}}, \bibinfo{pages}{6871} (\bibinfo{year}{2024}).

\bibitem{ye2024superanimal}
\bibinfo{author}{Ye, S.} \emph{et~al.}
\newblock \bibinfo{journal}{\bibinfo{title}{Superanimal pretrained pose estimation models for behavioral analysis}}.
\newblock {\emph{\JournalTitle{Nature Communications}}} \textbf{\bibinfo{volume}{15}}, \bibinfo{pages}{5165} (\bibinfo{year}{2024}).

\bibitem{pai2024foundation}
\bibinfo{author}{Pai, S.} \emph{et~al.}
\newblock \bibinfo{journal}{\bibinfo{title}{Foundation model for cancer imaging biomarkers}}.
\newblock {\emph{\JournalTitle{Nature machine intelligence}}} \textbf{\bibinfo{volume}{6}}, \bibinfo{pages}{354--367} (\bibinfo{year}{2024}).

\bibitem{cai2024pretrainable}
\bibinfo{author}{Cai, H.} \emph{et~al.}
\newblock \bibinfo{journal}{\bibinfo{title}{Pretrainable geometric graph neural network for antibody affinity maturation}}.
\newblock {\emph{\JournalTitle{Nature Communications}}} \textbf{\bibinfo{volume}{15}}, \bibinfo{pages}{7785} (\bibinfo{year}{2024}).

\bibitem{lu2022frozen}
\bibinfo{author}{Lu, K.}, \bibinfo{author}{Grover, A.}, \bibinfo{author}{Abbeel, P.} \& \bibinfo{author}{Mordatch, I.}
\newblock \bibinfo{title}{Frozen pretrained transformers as universal computation engines}.
\newblock In \emph{\bibinfo{booktitle}{Proceedings of the AAAI Conference on Artificial Intelligence}}, vol.~\bibinfo{volume}{36}, \bibinfo{pages}{7628--7636} (\bibinfo{year}{2022}).

\bibitem{dinh2022lift}
\bibinfo{author}{Dinh, T.} \emph{et~al.}
\newblock \bibinfo{journal}{\bibinfo{title}{Lift: Language-interfaced fine-tuning for non-language machine learning tasks}}.
\newblock {\emph{\JournalTitle{Advances in Neural Information Processing Systems}}} \textbf{\bibinfo{volume}{35}}, \bibinfo{pages}{11763--11784} (\bibinfo{year}{2022}).

\bibitem{vinod2023reprogramming}
\bibinfo{author}{Vinod, R.}, \bibinfo{author}{Chen, P.-Y.} \& \bibinfo{author}{Das, P.}
\newblock \bibinfo{journal}{\bibinfo{title}{Reprogramming pretrained language models for protein sequence representation learning}}.
\newblock {\emph{\JournalTitle{arXiv preprint arXiv:2301.02120}}}  (\bibinfo{year}{2023}).

\bibitem{ma2023medsam}
\bibinfo{author}{Ma, J.} \& \bibinfo{author}{Wang, B.}
\newblock \bibinfo{journal}{\bibinfo{title}{Segment anything in medical images}}.
\newblock {\emph{\JournalTitle{arXiv preprint arXiv:2304.12306}}}  (\bibinfo{year}{2023}).

\bibitem{shen2023cross}
\bibinfo{author}{Shen, J.} \emph{et~al.}
\newblock \bibinfo{title}{Cross-modal fine-tuning: Align then refine}.
\newblock In \emph{\bibinfo{booktitle}{International Conference on Machine Learning}}, \bibinfo{pages}{31030--31056} (\bibinfo{organization}{PMLR}, \bibinfo{year}{2023}).

\bibitem{hu2021lora}
\bibinfo{author}{Hu, E.~J.} \emph{et~al.}
\newblock \bibinfo{title}{Lora: Low-rank adaptation of large language models}.
\newblock In \emph{\bibinfo{booktitle}{International Conference on Learning Representations}} (\bibinfo{year}{2022}).

\bibitem{houlsby2019parameter}
\bibinfo{author}{Houlsby, N.} \emph{et~al.}
\newblock \bibinfo{title}{Parameter-efficient transfer learning for nlp}.
\newblock In \emph{\bibinfo{booktitle}{International conference on machine learning}}, \bibinfo{pages}{2790--2799} (\bibinfo{organization}{PMLR}, \bibinfo{year}{2019}).

\bibitem{zhu2023visual}
\bibinfo{author}{Zhu, J.}, \bibinfo{author}{Lai, S.}, \bibinfo{author}{Chen, X.}, \bibinfo{author}{Wang, D.} \& \bibinfo{author}{Lu, H.}
\newblock \bibinfo{title}{Visual prompt multi-modal tracking}.
\newblock In \emph{\bibinfo{booktitle}{Proceedings of the IEEE/CVF Conference on Computer Vision and Pattern Recognition}}, \bibinfo{pages}{9516--9526} (\bibinfo{year}{2023}).

\bibitem{glass_dataset}
\bibinfo{author}{Mei, H.} \emph{et~al.}
\newblock \bibinfo{title}{Glass segmentation using intensity and spectral polarization cues}.
\newblock In \emph{\bibinfo{booktitle}{Proceedings of the IEEE/CVF Conference on Computer Vision and Pattern Recognition}}, \bibinfo{pages}{12622--12631} (\bibinfo{year}{2022}).

\bibitem{zjurgbp}
\bibinfo{author}{Xiang, K.}, \bibinfo{author}{Yang, K.} \& \bibinfo{author}{Wang, K.}
\newblock \bibinfo{journal}{\bibinfo{title}{Polarization-driven semantic segmentation via efficient attention-bridged fusion}}.
\newblock {\emph{\JournalTitle{Optics Express}}} \textbf{\bibinfo{volume}{29}}, \bibinfo{pages}{4802--4820} (\bibinfo{year}{2021}).

\bibitem{Silberman:ECCV12}
\bibinfo{author}{Nathan~Silberman, P.~K., Derek~Hoiem} \& \bibinfo{author}{Fergus, R.}
\newblock \bibinfo{title}{Indoor segmentation and support inference from rgbd images}.
\newblock In \emph{\bibinfo{booktitle}{ECCV}} (\bibinfo{year}{2012}).

\bibitem{gupta2014learning}
\bibinfo{author}{Gupta, S.}, \bibinfo{author}{Girshick, R.}, \bibinfo{author}{Arbel{\'a}ez, P.} \& \bibinfo{author}{Malik, J.}
\newblock \bibinfo{title}{Learning rich features from rgb-d images for object detection and segmentation}.
\newblock In \emph{\bibinfo{booktitle}{Computer Vision--ECCV 2014: 13th European Conference, Zurich, Switzerland, September 6-12, 2014, Proceedings, Part VII 13}}, \bibinfo{pages}{345--360} (\bibinfo{organization}{Springer}, \bibinfo{year}{2014}).

\bibitem{huo2023glass}
\bibinfo{author}{Huo, D.}, \bibinfo{author}{Wang, J.}, \bibinfo{author}{Qian, Y.} \& \bibinfo{author}{Yang, Y.-H.}
\newblock \bibinfo{journal}{\bibinfo{title}{Glass segmentation with rgb-thermal image pairs}}.
\newblock {\emph{\JournalTitle{IEEE Transactions on Image Processing}}} \textbf{\bibinfo{volume}{32}}, \bibinfo{pages}{1911--1926} (\bibinfo{year}{2023}).

\bibitem{brown2011multi}
\bibinfo{author}{Brown, M.} \& \bibinfo{author}{S{\"u}sstrunk, S.}
\newblock \bibinfo{title}{Multi-spectral sift for scene category recognition}.
\newblock In \emph{\bibinfo{booktitle}{CVPR 2011}}, \bibinfo{pages}{177--184} (\bibinfo{organization}{IEEE}, \bibinfo{year}{2011}).

\bibitem{llava}
\bibinfo{author}{Liu, H.}, \bibinfo{author}{Li, C.}, \bibinfo{author}{Wu, Q.} \& \bibinfo{author}{Lee, Y.~J.}
\newblock \bibinfo{journal}{\bibinfo{title}{Visual instruction tuning}}.
\newblock {\emph{\JournalTitle{Advances in neural information processing systems}}} \textbf{\bibinfo{volume}{36}} (\bibinfo{year}{2024}).

\bibitem{li2023blip}
\bibinfo{author}{Li, J.}, \bibinfo{author}{Li, D.}, \bibinfo{author}{Savarese, S.} \& \bibinfo{author}{Hoi, S.}
\newblock \bibinfo{title}{Blip-2: Bootstrapping language-image pre-training with frozen image encoders and large language models}.
\newblock In \emph{\bibinfo{booktitle}{International conference on machine learning}}, \bibinfo{pages}{19730--19742} (\bibinfo{organization}{PMLR}, \bibinfo{year}{2023}).

\bibitem{sun2019rtfnet}
\bibinfo{author}{Sun, Y.}, \bibinfo{author}{Zuo, W.} \& \bibinfo{author}{Liu, M.}
\newblock \bibinfo{journal}{\bibinfo{title}{Rtfnet: Rgb-thermal fusion network for semantic segmentation of urban scenes}}.
\newblock {\emph{\JournalTitle{IEEE Robotics and Automation Letters}}} \textbf{\bibinfo{volume}{4}}, \bibinfo{pages}{2576--2583} (\bibinfo{year}{2019}).

\bibitem{singh2023depth}
\bibinfo{author}{Singh, A.~D.} \emph{et~al.}
\newblock \bibinfo{title}{Depth estimation from camera image and mmwave radar point cloud}.
\newblock In \emph{\bibinfo{booktitle}{Proceedings of the IEEE/CVF Conference on Computer Vision and Pattern Recognition}}, \bibinfo{pages}{9275--9285} (\bibinfo{year}{2023}).

\bibitem{chen2020simple}
\bibinfo{author}{Chen, T.}, \bibinfo{author}{Kornblith, S.}, \bibinfo{author}{Norouzi, M.} \& \bibinfo{author}{Hinton, G.}
\newblock \bibinfo{title}{A simple framework for contrastive learning of visual representations}.
\newblock In \emph{\bibinfo{booktitle}{International conference on machine learning}}, \bibinfo{pages}{1597--1607} (\bibinfo{organization}{PMLR}, \bibinfo{year}{2020}).

\bibitem{moco}
\bibinfo{author}{He, K.}, \bibinfo{author}{Fan, H.}, \bibinfo{author}{Wu, Y.}, \bibinfo{author}{Xie, S.} \& \bibinfo{author}{Girshick, R.}
\newblock \bibinfo{title}{Momentum contrast for unsupervised visual representation learning}.
\newblock In \emph{\bibinfo{booktitle}{Proceedings of the IEEE/CVF conference on computer vision and pattern recognition}}, \bibinfo{pages}{9729--9738} (\bibinfo{year}{2020}).

\bibitem{jia2022visual}
\bibinfo{author}{Jia, M.} \emph{et~al.}
\newblock \bibinfo{title}{Visual prompt tuning}.
\newblock In \emph{\bibinfo{booktitle}{European Conference on Computer Vision}}, \bibinfo{pages}{709--727} (\bibinfo{organization}{Springer}, \bibinfo{year}{2022}).

\bibitem{ouyang2022training}
\bibinfo{author}{Ouyang, L.} \emph{et~al.}
\newblock \bibinfo{journal}{\bibinfo{title}{Training language models to follow instructions with human feedback}}.
\newblock {\emph{\JournalTitle{Advances in neural information processing systems}}} \textbf{\bibinfo{volume}{35}}, \bibinfo{pages}{27730--27744} (\bibinfo{year}{2022}).

\bibitem{clip}
\bibinfo{author}{Radford, A.} \emph{et~al.}
\newblock \bibinfo{title}{Learning transferable visual models from natural language supervision}.
\newblock In \emph{\bibinfo{booktitle}{International conference on machine learning}}, \bibinfo{pages}{8748--8763} (\bibinfo{organization}{PMLR}, \bibinfo{year}{2021}).

\bibitem{controlnet}
\bibinfo{author}{Zhang, L.}, \bibinfo{author}{Rao, A.} \& \bibinfo{author}{Agrawala, M.}
\newblock \bibinfo{title}{Adding conditional control to text-to-image diffusion models}.
\newblock In \emph{\bibinfo{booktitle}{Proceedings of the IEEE/CVF International Conference on Computer Vision}}, \bibinfo{pages}{3836--3847} (\bibinfo{year}{2023}).

\bibitem{lisa}
\bibinfo{author}{Lai, X.} \emph{et~al.}
\newblock \bibinfo{title}{Lisa: Reasoning segmentation via large language model}.
\newblock In \emph{\bibinfo{booktitle}{Proceedings of the IEEE/CVF Conference on Computer Vision and Pattern Recognition}}, \bibinfo{pages}{9579--9589} (\bibinfo{year}{2024}).

\bibitem{rombach2021highresolution}
\bibinfo{author}{Rombach, R.}, \bibinfo{author}{Blattmann, A.}, \bibinfo{author}{Lorenz, D.}, \bibinfo{author}{Esser, P.} \& \bibinfo{author}{Ommer, B.}
\newblock \bibinfo{title}{High-resolution image synthesis with latent diffusion models} (\bibinfo{year}{2021}).
\newblock \eprint{2112.10752}.

\bibitem{zamir2018taskonomy}
\bibinfo{author}{Zamir, A.~R.} \emph{et~al.}
\newblock \bibinfo{title}{Taskonomy: Disentangling task transfer learning}.
\newblock In \emph{\bibinfo{booktitle}{Proceedings of the IEEE conference on computer vision and pattern recognition}} (\bibinfo{year}{2018}).

\bibitem{deng2009imagenet}
\bibinfo{author}{Deng, J.} \emph{et~al.}
\newblock \bibinfo{title}{Imagenet: A large-scale hierarchical image database}.
\newblock In \emph{\bibinfo{booktitle}{2009 IEEE conference on computer vision and pattern recognition}}, \bibinfo{pages}{248--255} (\bibinfo{organization}{Ieee}, \bibinfo{year}{2009}).

\bibitem{moco-v2}
\bibinfo{author}{Chen, X.}, \bibinfo{author}{Fan, H.}, \bibinfo{author}{Girshick, R.} \& \bibinfo{author}{He, K.}
\newblock \bibinfo{journal}{\bibinfo{title}{Improved baselines with momentum contrastive learning}}.
\newblock {\emph{\JournalTitle{arXiv preprint arXiv:2003.04297}}}  (\bibinfo{year}{2020}).

\bibitem{moco-v3}
\bibinfo{author}{Fan, H.} \emph{et~al.}
\newblock \bibinfo{title}{Multiscale vision transformers}.
\newblock In \emph{\bibinfo{booktitle}{Proceedings of the IEEE/CVF International Conference on Computer Vision (ICCV)}}, \bibinfo{pages}{6824--6835} (\bibinfo{year}{2021}).

\bibitem{BYOL}
\bibinfo{author}{Grill, J.-B.} \emph{et~al.}
\newblock \bibinfo{journal}{\bibinfo{title}{Bootstrap your own latent-a new approach to self-supervised learning}}.
\newblock {\emph{\JournalTitle{Advances in neural information processing systems}}} \textbf{\bibinfo{volume}{33}}, \bibinfo{pages}{21271--21284} (\bibinfo{year}{2020}).

\bibitem{he2022masked}
\bibinfo{author}{He, K.} \emph{et~al.}
\newblock \bibinfo{title}{Masked autoencoders are scalable vision learners}.
\newblock In \emph{\bibinfo{booktitle}{Proceedings of the IEEE/CVF conference on computer vision and pattern recognition}}, \bibinfo{pages}{16000--16009} (\bibinfo{year}{2022}).

\bibitem{simmim}
\bibinfo{author}{Xie, Z.} \emph{et~al.}
\newblock \bibinfo{title}{Simmim: A simple framework for masked image modeling}.
\newblock In \emph{\bibinfo{booktitle}{Proceedings of the IEEE/CVF Conference on Computer Vision and Pattern Recognition}}, \bibinfo{pages}{9653--9663} (\bibinfo{year}{2022}).

\bibitem{beit}
\bibinfo{author}{Bao, H.}, \bibinfo{author}{Dong, L.}, \bibinfo{author}{Piao, S.} \& \bibinfo{author}{Wei, F.}
\newblock \bibinfo{title}{Beit: Bert pre-training of image transformers}.
\newblock In \emph{\bibinfo{booktitle}{International Conference on Learning Representations}} (\bibinfo{year}{2022}).

\bibitem{singh2019rna}
\bibinfo{author}{Singh, J.}, \bibinfo{author}{Hanson, J.}, \bibinfo{author}{Paliwal, K.} \& \bibinfo{author}{Zhou, Y.}
\newblock \bibinfo{journal}{\bibinfo{title}{Rna secondary structure prediction using an ensemble of two-dimensional deep neural networks and transfer learning}}.
\newblock {\emph{\JournalTitle{Nature communications}}} \textbf{\bibinfo{volume}{10}}, \bibinfo{pages}{5407} (\bibinfo{year}{2019}).

\bibitem{kang2023multi}
\bibinfo{author}{Kang, Y.}, \bibinfo{author}{Park, H.}, \bibinfo{author}{Smit, B.} \& \bibinfo{author}{Kim, J.}
\newblock \bibinfo{journal}{\bibinfo{title}{A multi-modal pre-training transformer for universal transfer learning in metal--organic frameworks}}.
\newblock {\emph{\JournalTitle{Nature Machine Intelligence}}} \textbf{\bibinfo{volume}{5}}, \bibinfo{pages}{309--318} (\bibinfo{year}{2023}).

\bibitem{pan2023transfer}
\bibinfo{author}{Pan, J.}
\newblock \bibinfo{journal}{\bibinfo{title}{Transfer learning for metal--organic frameworks}}.
\newblock {\emph{\JournalTitle{Nature Computational Science}}} \textbf{\bibinfo{volume}{3}}, \bibinfo{pages}{280--280} (\bibinfo{year}{2023}).

\bibitem{wang2021predicting}
\bibinfo{author}{Wang, K.}, \bibinfo{author}{Johnson, C.~W.}, \bibinfo{author}{Bennett, K.~C.} \& \bibinfo{author}{Johnson, P.~A.}
\newblock \bibinfo{journal}{\bibinfo{title}{Predicting fault slip via transfer learning}}.
\newblock {\emph{\JournalTitle{Nature communications}}} \textbf{\bibinfo{volume}{12}}, \bibinfo{pages}{7319} (\bibinfo{year}{2021}).

\bibitem{choi2023neural}
\bibinfo{author}{Choi, E.} \emph{et~al.}
\newblock \bibinfo{journal}{\bibinfo{title}{Neural 360 structured light with learned metasurfaces}}.
\newblock {\emph{\JournalTitle{arXiv preprint arXiv:2306.13361}}}  (\bibinfo{year}{2023}).

\bibitem{guo2024eventlfm}
\bibinfo{author}{Guo, R.} \emph{et~al.}
\newblock \bibinfo{journal}{\bibinfo{title}{Eventlfm: Event camera integrated fourier light field microscopy for ultrafast 3d imaging}}.
\newblock {\emph{\JournalTitle{Light: Science \& Applications}}} \textbf{\bibinfo{volume}{13}}, \bibinfo{pages}{144} (\bibinfo{year}{2024}).

\bibitem{radhakrishnan2023transfer}
\bibinfo{author}{Radhakrishnan, A.}, \bibinfo{author}{Ruiz~Luyten, M.}, \bibinfo{author}{Prasad, N.} \& \bibinfo{author}{Uhler, C.}
\newblock \bibinfo{journal}{\bibinfo{title}{Transfer learning with kernel methods}}.
\newblock {\emph{\JournalTitle{Nature Communications}}} \textbf{\bibinfo{volume}{14}}, \bibinfo{pages}{5570} (\bibinfo{year}{2023}).

\bibitem{li2024llava}
\bibinfo{author}{Li, B.} \emph{et~al.}
\newblock \bibinfo{journal}{\bibinfo{title}{Llava-onevision: Easy visual task transfer}}.
\newblock {\emph{\JournalTitle{arXiv preprint arXiv:2408.03326}}}  (\bibinfo{year}{2024}).

\bibitem{zhang2024generalist}
\bibinfo{author}{Zhang, K.} \emph{et~al.}
\newblock \bibinfo{journal}{\bibinfo{title}{A generalist vision--language foundation model for diverse biomedical tasks}}.
\newblock {\emph{\JournalTitle{Nature Medicine}}} \bibinfo{pages}{1--13} (\bibinfo{year}{2024}).

\bibitem{da_adverse_learning_1}
\bibinfo{author}{Vu, T.-H.}, \bibinfo{author}{Jain, H.}, \bibinfo{author}{Bucher, M.}, \bibinfo{author}{Cord, M.} \& \bibinfo{author}{P{\'e}rez, P.}
\newblock \bibinfo{title}{Advent: Adversarial entropy minimization for domain adaptation in semantic segmentation}.
\newblock In \emph{\bibinfo{booktitle}{Proceedings of the IEEE/CVF conference on computer vision and pattern recognition}}, \bibinfo{pages}{2517--2526} (\bibinfo{year}{2019}).

\bibitem{da_adverse_learning_2}
\bibinfo{author}{Pan, F.}, \bibinfo{author}{Shin, I.}, \bibinfo{author}{Rameau, F.}, \bibinfo{author}{Lee, S.} \& \bibinfo{author}{Kweon, I.~S.}
\newblock \bibinfo{title}{Unsupervised intra-domain adaptation for semantic segmentation through self-supervision}.
\newblock In \emph{\bibinfo{booktitle}{Proceedings of the IEEE/CVF Conference on Computer Vision and Pattern Recognition}}, \bibinfo{pages}{3764--3773} (\bibinfo{year}{2020}).

\bibitem{da_adverse_learning_3}
\bibinfo{author}{Lai, X.} \emph{et~al.}
\newblock \bibinfo{title}{Decouplenet: Decoupled network for domain adaptive semantic segmentation}.
\newblock In \emph{\bibinfo{booktitle}{European Conference on Computer Vision}}, \bibinfo{pages}{369--387} (\bibinfo{organization}{Springer}, \bibinfo{year}{2022}).

\bibitem{da_self_learning_1}
\bibinfo{author}{Chen, M.}, \bibinfo{author}{Xue, H.} \& \bibinfo{author}{Cai, D.}
\newblock \bibinfo{title}{Domain adaptation for semantic segmentation with maximum squares loss}.
\newblock In \emph{\bibinfo{booktitle}{Proceedings of the IEEE/CVF International Conference on Computer Vision}}, \bibinfo{pages}{2090--2099} (\bibinfo{year}{2019}).

\bibitem{da_self_learning_2}
\bibinfo{author}{Zou, Y.}, \bibinfo{author}{Yu, Z.}, \bibinfo{author}{Kumar, B.} \& \bibinfo{author}{Wang, J.}
\newblock \bibinfo{title}{Unsupervised domain adaptation for semantic segmentation via class-balanced self-training}.
\newblock In \emph{\bibinfo{booktitle}{Proceedings of the European conference on computer vision (ECCV)}}, \bibinfo{pages}{289--305} (\bibinfo{year}{2018}).

\bibitem{da_self_learning_3}
\bibinfo{author}{Wang, P.} \emph{et~al.}
\newblock \bibinfo{journal}{\bibinfo{title}{Uncertainty-aware clustering for unsupervised domain adaptive object re-identification}}.
\newblock {\emph{\JournalTitle{IEEE Transactions on Multimedia}}}  (\bibinfo{year}{2022}).

\bibitem{hda_1}
\bibinfo{author}{Liu, F.}, \bibinfo{author}{Zhang, G.} \& \bibinfo{author}{Lu, J.}
\newblock \bibinfo{journal}{\bibinfo{title}{Heterogeneous domain adaptation: An unsupervised approach}}.
\newblock {\emph{\JournalTitle{IEEE transactions on neural networks and learning systems}}} \textbf{\bibinfo{volume}{31}}, \bibinfo{pages}{5588--5602} (\bibinfo{year}{2020}).

\bibitem{hda_2}
\bibinfo{author}{Luo, Y.}, \bibinfo{author}{Wen, Y.}, \bibinfo{author}{Liu, T.} \& \bibinfo{author}{Tao, D.}
\newblock \bibinfo{journal}{\bibinfo{title}{Transferring knowledge fragments for learning distance metric from a heterogeneous domain}}.
\newblock {\emph{\JournalTitle{IEEE transactions on pattern analysis and machine intelligence}}} \textbf{\bibinfo{volume}{41}}, \bibinfo{pages}{1013--1026} (\bibinfo{year}{2018}).

\bibitem{chen2022adaptformer}
\bibinfo{author}{Chen, S.} \emph{et~al.}
\newblock \bibinfo{journal}{\bibinfo{title}{Adaptformer: Adapting vision transformers for scalable visual recognition}}.
\newblock {\emph{\JournalTitle{Advances in Neural Information Processing Systems}}} \textbf{\bibinfo{volume}{35}}, \bibinfo{pages}{16664--16678} (\bibinfo{year}{2022}).

\bibitem{he2016deep}
\bibinfo{author}{He, K.}, \bibinfo{author}{Zhang, X.}, \bibinfo{author}{Ren, S.} \& \bibinfo{author}{Sun, J.}
\newblock \bibinfo{title}{Deep residual learning for image recognition}.
\newblock In \emph{\bibinfo{booktitle}{Proceedings of the IEEE conference on computer vision and pattern recognition}}, \bibinfo{pages}{770--778} (\bibinfo{year}{2016}).

\bibitem{li2021prefix}
\bibinfo{author}{Li, X.~L.} \& \bibinfo{author}{Liang, P.}
\newblock \bibinfo{journal}{\bibinfo{title}{Prefix-tuning: Optimizing continuous prompts for generation}}.
\newblock {\emph{\JournalTitle{arXiv preprint arXiv:2101.00190}}}  (\bibinfo{year}{2021}).

\bibitem{internimage}
\bibinfo{author}{Wang, W.} \emph{et~al.}
\newblock \bibinfo{title}{Internimage: Exploring large-scale vision foundation models with deformable convolutions}.
\newblock In \emph{\bibinfo{booktitle}{Proceedings of the IEEE/CVF Conference on Computer Vision and Pattern Recognition}}, \bibinfo{pages}{14408--14419} (\bibinfo{year}{2023}).

\bibitem{imagebind}
\bibinfo{author}{Girdhar, R.} \emph{et~al.}
\newblock \bibinfo{title}{Imagebind: One embedding space to bind them all}.
\newblock In \emph{\bibinfo{booktitle}{Proceedings of the IEEE/CVF Conference on Computer Vision and Pattern Recognition}}, \bibinfo{pages}{15180--15190} (\bibinfo{year}{2023}).

\bibitem{vaswani2017attention}
\bibinfo{author}{Vaswani, A.} \emph{et~al.}
\newblock \bibinfo{journal}{\bibinfo{title}{Attention is all you need}}.
\newblock {\emph{\JournalTitle{Advances in neural information processing systems}}} \textbf{\bibinfo{volume}{30}} (\bibinfo{year}{2017}).

\bibitem{touvron2023llama}
\bibinfo{author}{Touvron, H.} \emph{et~al.}
\newblock \bibinfo{journal}{\bibinfo{title}{Llama 2: Open foundation and fine-tuned chat models}}.
\newblock {\emph{\JournalTitle{arXiv preprint arXiv:2307.09288}}}  (\bibinfo{year}{2023}).

\bibitem{dosovitskiy2020image}
\bibinfo{author}{Dosovitskiy, A.} \emph{et~al.}
\newblock \bibinfo{title}{An image is worth 16x16 words: Transformers for image recognition at scale}.
\newblock In \emph{\bibinfo{booktitle}{International Conference on Learning Representations}} (\bibinfo{year}{2021}).

\bibitem{liu2021swin}
\bibinfo{author}{Liu, Z.} \emph{et~al.}
\newblock \bibinfo{title}{Swin transformer: Hierarchical vision transformer using shifted windows}.
\newblock In \emph{\bibinfo{booktitle}{Proceedings of the IEEE/CVF international conference on computer vision}}, \bibinfo{pages}{10012--10022} (\bibinfo{year}{2021}).

\bibitem{wang2021not}
\bibinfo{author}{Wang, Y.}, \bibinfo{author}{Huang, R.}, \bibinfo{author}{Song, S.}, \bibinfo{author}{Huang, Z.} \& \bibinfo{author}{Huang, G.}
\newblock \bibinfo{journal}{\bibinfo{title}{Not all images are worth 16x16 words: Dynamic transformers for efficient image recognition}}.
\newblock {\emph{\JournalTitle{Advances in Neural Information Processing Systems}}} \textbf{\bibinfo{volume}{34}}, \bibinfo{pages}{11960--11973} (\bibinfo{year}{2021}).

\bibitem{guo2021pct}
\bibinfo{author}{Guo, M.-H.} \emph{et~al.}
\newblock \bibinfo{journal}{\bibinfo{title}{Pct: Point cloud transformer}}.
\newblock {\emph{\JournalTitle{Computational Visual Media}}} \textbf{\bibinfo{volume}{7}}, \bibinfo{pages}{187--199} (\bibinfo{year}{2021}).

\bibitem{zhao2021point}
\bibinfo{author}{Zhao, H.}, \bibinfo{author}{Jiang, L.}, \bibinfo{author}{Jia, J.}, \bibinfo{author}{Torr, P.~H.} \& \bibinfo{author}{Koltun, V.}
\newblock \bibinfo{title}{Point transformer}.
\newblock In \emph{\bibinfo{booktitle}{Proceedings of the IEEE/CVF international conference on computer vision}}, \bibinfo{pages}{16259--16268} (\bibinfo{year}{2021}).

\bibitem{wu2022point}
\bibinfo{author}{Wu, X.}, \bibinfo{author}{Lao, Y.}, \bibinfo{author}{Jiang, L.}, \bibinfo{author}{Liu, X.} \& \bibinfo{author}{Zhao, H.}
\newblock \bibinfo{journal}{\bibinfo{title}{Point transformer v2: Grouped vector attention and partition-based pooling}}.
\newblock {\emph{\JournalTitle{Advances in Neural Information Processing Systems}}} \textbf{\bibinfo{volume}{35}}, \bibinfo{pages}{33330--33342} (\bibinfo{year}{2022}).

\bibitem{gong2021audio}
\bibinfo{author}{Gong, Y.}, \bibinfo{author}{Chung, Y.-A.} \& \bibinfo{author}{Glass, J.}
\newblock \bibinfo{journal}{\bibinfo{title}{Ast: Audio spectrogram transformer}}.
\newblock {\emph{\JournalTitle{arXiv preprint arXiv:2104.01778}}}  (\bibinfo{year}{2021}).

\bibitem{chen2022hts}
\bibinfo{author}{Chen, K.} \emph{et~al.}
\newblock \bibinfo{title}{Hts-at: A hierarchical token-semantic audio transformer for sound classification and detection}.
\newblock In \emph{\bibinfo{booktitle}{ICASSP 2022-2022 IEEE International Conference on Acoustics, Speech and Signal Processing (ICASSP)}}, \bibinfo{pages}{646--650} (\bibinfo{organization}{IEEE}, \bibinfo{year}{2022}).

\bibitem{verma2021audio}
\bibinfo{author}{Verma, P.} \& \bibinfo{author}{Berger, J.}
\newblock \bibinfo{journal}{\bibinfo{title}{Audio transformers: Transformer architectures for large scale audio understanding. adieu convolutions}}.
\newblock {\emph{\JournalTitle{arXiv preprint arXiv:2105.00335}}}  (\bibinfo{year}{2021}).

\bibitem{jaegle2021perceiver}
\bibinfo{author}{Jaegle, A.} \emph{et~al.}
\newblock \bibinfo{title}{Perceiver: General perception with iterative attention}.
\newblock In \emph{\bibinfo{booktitle}{International conference on machine learning}}, \bibinfo{pages}{4651--4664} (\bibinfo{organization}{PMLR}, \bibinfo{year}{2021}).

\bibitem{jaegle2021perceiverio}
\bibinfo{author}{Jaegle, A.} \emph{et~al.}
\newblock \bibinfo{title}{Perceiver io: A general architecture for structured inputs \& outputs}.
\newblock In \emph{\bibinfo{booktitle}{International Conference on Learning Representations}} (\bibinfo{year}{2022}).

\bibitem{tamkin2021dabs}
\bibinfo{author}{Tamkin, A.} \emph{et~al.}
\newblock \bibinfo{journal}{\bibinfo{title}{Dabs: A domain-agnostic benchmark for self-supervised learning}}.
\newblock {\emph{\JournalTitle{Advances in neural information processing systems}}}  (\bibinfo{year}{2021}).

\bibitem{wu2023randomized}
\bibinfo{author}{Wu, H.} \emph{et~al.}
\newblock \bibinfo{title}{Randomized quantization: A generic augmentation for data agnostic self-supervised learning}.
\newblock In \emph{\bibinfo{booktitle}{Proceedings of the IEEE/CVF International Conference on Computer Vision}}, \bibinfo{pages}{16305--16316} (\bibinfo{year}{2023}).

\bibitem{wang2023internimage}
\bibinfo{author}{Wang, W.} \emph{et~al.}
\newblock \bibinfo{title}{Internimage: Exploring large-scale vision foundation models with deformable convolutions}.
\newblock In \emph{\bibinfo{booktitle}{Proceedings of the IEEE/CVF conference on computer vision and pattern recognition}}, \bibinfo{pages}{14408--14419} (\bibinfo{year}{2023}).

\bibitem{zhang2023meta}
\bibinfo{author}{Zhang, Y.} \emph{et~al.}
\newblock \bibinfo{journal}{\bibinfo{title}{Meta-transformer: A unified framework for multimodal learning}}.
\newblock {\emph{\JournalTitle{arXiv preprint arXiv:2307.10802}}}  (\bibinfo{year}{2023}).

\bibitem{xdecoder}
\bibinfo{author}{Zou, X.} \emph{et~al.}
\newblock \bibinfo{title}{Generalized decoding for pixel, image, and language}.
\newblock In \emph{\bibinfo{booktitle}{Proceedings of the IEEE/CVF Conference on Computer Vision and Pattern Recognition}}, \bibinfo{pages}{15116--15127} (\bibinfo{year}{2023}).

\bibitem{4m}
\bibinfo{author}{Mizrahi, D.} \emph{et~al.}
\newblock \bibinfo{journal}{\bibinfo{title}{4m: Massively multimodal masked modeling}}.
\newblock {\emph{\JournalTitle{Advances in Neural Information Processing Systems}}} \textbf{\bibinfo{volume}{36}} (\bibinfo{year}{2024}).

\bibitem{seem}
\bibinfo{author}{Zou, X.} \emph{et~al.}
\newblock \bibinfo{title}{Segment everything everywhere all at once}.
\newblock In \emph{\bibinfo{booktitle}{NeurIPS}} (\bibinfo{year}{2023}).

\bibitem{ji2023sam}
\bibinfo{author}{Ji, G.-P.} \emph{et~al.}
\newblock \bibinfo{journal}{\bibinfo{title}{Sam struggles in concealed scenes--empirical study on" segment anything"}}.
\newblock {\emph{\JournalTitle{arXiv preprint arXiv:2304.06022}}}  (\bibinfo{year}{2023}).

\bibitem{tang2023camouflage}
\bibinfo{author}{Tang, L.}, \bibinfo{author}{Xiao, H.} \& \bibinfo{author}{Li, B.}
\newblock \bibinfo{journal}{\bibinfo{title}{Can sam segment anything? when sam meets camouflaged object detection}}.
\newblock {\emph{\JournalTitle{arXiv preprint arXiv:2304.04709}}}  (\bibinfo{year}{2023}).

\bibitem{chen2023samfail}
\bibinfo{author}{Chen, T.} \emph{et~al.}
\newblock \bibinfo{journal}{\bibinfo{title}{Sam fails to segment anything?--sam-adapter: Adapting sam in underperformed scenes: Camouflage, shadow, and more}}.
\newblock {\emph{\JournalTitle{arXiv preprint arXiv:2304.09148}}}  (\bibinfo{year}{2023}).

\bibitem{cen2023sad}
\bibinfo{author}{Cen, J.} \emph{et~al.}
\newblock \bibinfo{journal}{\bibinfo{title}{Sad: Segment any rgbd}}.
\newblock {\emph{\JournalTitle{arXiv preprint arXiv:2305.14207}}}  (\bibinfo{year}{2023}).

\bibitem{zhang2023cmx}
\bibinfo{author}{Zhang, J.} \emph{et~al.}
\newblock \bibinfo{journal}{\bibinfo{title}{Cmx: Cross-modal fusion for rgb-x semantic segmentation with transformers}}.
\newblock {\emph{\JournalTitle{IEEE Transactions on Intelligent Transportation Systems}}}  (\bibinfo{year}{2023}).

\bibitem{he2021towards}
\bibinfo{author}{He, J.}, \bibinfo{author}{Zhou, C.}, \bibinfo{author}{Ma, X.}, \bibinfo{author}{Berg-Kirkpatrick, T.} \& \bibinfo{author}{Neubig, G.}
\newblock \bibinfo{journal}{\bibinfo{title}{Towards a unified view of parameter-efficient transfer learning}}.
\newblock {\emph{\JournalTitle{arXiv preprint arXiv:2110.04366}}}  (\bibinfo{year}{2021}).

\end{thebibliography}
\clearpage

\begin{table}[h!]
\centering
\resizebox{\textwidth}{!}{  
\begin{tabular}{>{\columncolor{gray!30}}l>{\columncolor{gray!30}}l>{\columncolor{gray!30}}l>{\columncolor{gray!30}}l}
\hline
\rowcolor{gray!30}
\textbf{Choice of MAT} & \textbf{Required data} & \textbf{Efficiency} & \textbf{Performance (\% mIoU)} \\
\hline
Training from scratch & \cellcolor{green!20}Target data only & \cellcolor{green!20}Efficient & \cellcolor{green!20}25.43 (Bad performance)  \\
\hline
Querying transformer in BLIP-2~\cite{li2023blip} & \cellcolor{red!20}Source-target pair & \cellcolor{red!20}NA (Not applicable) & \cellcolor{red!20}NA (Not applicable) \\
\hline
Dimension alignment in ORCA~\cite{shen2023cross} & \cellcolor{red!20}Source-target pair & \cellcolor{red!20}NA (Not applicable) & \cellcolor{red!20}NA (Not applicable) \\
\hline
Randomly Initialized Patch Embedding~\cite{lu2022frozen} & \cellcolor{green!20}Target data only & \cellcolor{green!20}Efficient & \cellcolor{red!20}58.89 (Bad performance) \\
\hline
Linear Layer + Trainable Patch Embedding~\cite{llava} & \cellcolor{green!20}Target data only & \cellcolor{green!20}Efficient & \cellcolor{red!20}63.96 (Moderate Performance) \\
\hline
Transformers + Trainable Patch Embedding~\cite{zhang2023cmx} & \cellcolor{green!20}Target data only & \cellcolor{red!20}More parameters & \cellcolor{red!20}26.67 (Bad performance) \\
\hline
Transpose to Batch Dimension~\cite{wu2023medical} & \cellcolor{green!20}Target data only & \cellcolor{red!20}Much more FLOPS & \cellcolor{green!20}70.90 (Good performance) \\
\hline
Ours & \cellcolor{green!20}Target data only & \cellcolor{green!20}Efficient & \cellcolor{green!20}72.69 (Best performance) \\
\hline
\end{tabular}
}
\caption{\textbf{Exploring Modality-agnostic Transfer Layers. } Randomly initializing a patch embedding for each modality leads to worse results than any method that inherits the vision embedding layer. A simple $1 \times 1$ convolution can improve the performance already. Interestingly, when the pretrained vision embedding layer is used, the performance would be better if we fixed the weights. All models are trained with Adapter finetuning strategy.}
\label{table:MAT_design}
\end{table}

\clearpage

\noindent This supplementary document provides additional descriptions and results to support the findings from the main manuscript.

\section{Additional Training Details}
\label{sec:add_training}
We report the effect of different finetuning strategies on trainable parameters in Table~\ref{tab:params}. The foundation model SAM with ViT-B~\cite{dosovitskiy2020image} as backbone contains 93.7M parameters from the image encoder, prompt encoder, and mask decoder. 
Full finetuning makes all parameters trainable. 
For parameter-efficient tuning, we implement four typical methods including LoRA~\cite{hu2021lora}, MLP adapter~\cite{houlsby2019parameter}, and prompt tuning~\cite{jia2022visual}. Following He et al.~\cite{he2021towards}, we balance their trainable parameters to achieve approximately 4\% of full parameters for fair comparison.

The detailed training configuration is presented in Table~\ref{tab:config}. 
We fix the training epoch to 50 and set the batch size as 4 regardless of the number of training samples in different modality datasets.
We sweep the learning rates from 3e-6 to 3e-3 and report the peak performance as the final result. 
The input modality images are resized to (1024, 1024) to meet the requirements of SAM.

\begin{table}[h]
\begin{center}
\renewcommand{\arraystretch}{1.1} %
\resizebox{0.6\linewidth}{!}{
\begin{tabular}{l|c}
    \specialrule{1pt}{0pt}{0pt}
      Finetuning Strategies & Trainable Parameters (M) \\
       & of Foundation Model  \\
       \hline
    LoRA &  4.3  \\ 
    MLP adapter & 3.9 \\
    Prompt tuning & 4.4 \\
    Full finetuning & 93.7 \\
    \specialrule{1pt}{0pt}{0pt}
    
\end{tabular}
}
\end{center}
\caption{ \textbf{The Number of Trainable Parameters in Foundation Model (SAM) with Different Finetuing Strategies.} Three parameter-efficient finetuning methods hold similar trainable parameters, which are much less than the trainable parameters of full finetuning strategies.}
\label{tab:params}
\end{table}

\begin{table}[h]
\begin{center}
\renewcommand{\arraystretch}{1.1} %
\resizebox{0.7\linewidth}{!}{
\begin{tabular}{l|c}
    \specialrule{1pt}{0pt}{0pt}
      Config & Value \\
     \hline 
     optimizer  & Adam  \\
     optimizer momentum &  $\beta_1$,$\beta_2$=0.9,0.999 \\
     batch size & 4 \\
     epoch & 50 \\ 
     learning rate & \{3e-6, 1e-5, 3e-5, 1e-4, 3e-4, 1e-3, 3e-3\} \\
     learning rate schedule & step decay \\
     schedule step size & 10 epoch \\
     schedule gamma & 0.5\\
     augmentation & \text{Resize}(1024, 1024) \\
      \specialrule{1pt}{0pt}{0pt}
    
\end{tabular}
}
\end{center}
\caption{ \textbf{The Training Setting for Our Experiments.} }
\label{tab:config}
\end{table}

\clearpage

\section{Additional Controlled Experiments}
We provide the study of the hyper-parameter setting of SimMAT by applying it to the Polarization modality. As shown in Figure~\ref{fig:effect}, SimMAT stack the \textit{n} convolutional layers with \textit{k} kernel size and dimension \textit{d}. SimMAT achieves best 72.7\% mIoU by setting {\textit{n}, \textit{k}, \textit{d}\} as  \{2, 3, 64\}. Further increasing the number of stacked layers and dimensional does not bring additional improvements, we suspect it is caused by the factor that introducing more trainable parameters makes training of SimMAT more challenging.
Note that when the kernel size is set to 1 and layers are set to 2, the implementation is the same as an MLP layer adopted in contrastive learning~\cite{chen2020simple,moco}. When the kernel size is set to 1 and layers are set to 1, the implementation is the same as a simple linear layer. One can observe that setting kernel size to 3 achieves peak performance with the best tradeoff between the receptive field and trainable parameters.

\begin{figure*}[h!]
\centering
\begin{minipage}{0.3\textwidth}
\centering
\resizebox{1\textwidth}{!}{
\begin{tabular}{l|ccc}
\hline
\textit{k} & 1 & 3 & 5  \\
\hline
mIoU(\%) & 71.7 & 72.7 & 71.3 \\
\hline
\end{tabular}
}
\end{minipage}
\hfill
\begin{minipage}{0.3\textwidth}
\centering
\resizebox{1\textwidth}{!}{
\begin{tabular}{l|ccc}
\hline
\textit{d} & 32 & 64 & 96 \\
\hline
mIoU(\%) & 71.5 & 72.7 & 72.7 \\
\hline
\end{tabular}
}
\end{minipage}
\hfill
\begin{minipage}{0.35\textwidth}
\centering
\resizebox{1\textwidth}{!}{
\begin{tabular}{l|ccccc}
\hline
\textit{n} & 1 & 2 & 3 & 4 & 5 \\
\hline
mIoU(\%) & 69.8 & 72.7 & 71.1 & 71.3 & 71.8 \\
\hline
Params(K) &  0.03 & 5.4 & 42.3 & 79.3 & 116.2 \\ 
\hline
\end{tabular}
}
\end{minipage}%
\vspace{1mm}

\begin{minipage}{0.3\textwidth}
\centering
    The effect of kernel size.
\end{minipage}
\begin{minipage}{0.3\textwidth}
\centering
    The effect of dimension.
\end{minipage}
\begin{minipage}{0.35\textwidth}
\centering
    The effect of layers.
\end{minipage}

\caption{\textbf{The Effect of the Configuration of our MAT layers, evaluated on Polarization modality.} Based on the above results, we set the $k,d,n$ to 3, 64, and 2, respectively, considering the trade-off of performance of efficiency.}
\label{fig:effect}
\end{figure*}

\clearpage

\section{Additional Comparisons}
\label{sec:add_baselines}
We report the training curve of SimMAT and baselines on the Polarization dataset in Figure~\ref{fig:supp_curve}. 
One can observe the training from scratch only achieves 25.43\% mIoU, significantly worse than other methods using prestrained weight as initialization. 
To tackle the channel misalignment between RGB modality and new modality input, two straightforward ideas are to build a new randomly initialized patch embedding or prepend a 1$\times$1 convolution layer for dimension projection. While these two methods achieve significant improvement over training from scratch, {their performance is suboptimal.} Our SimMAT achieves a better performance over these two commonly adopted naive baselines. 

Besides, we compare our SimMAT to two SOTA methods with pseudo new modality (RGBX) input. 
\textbf{ViPT}~\cite{zhu2023visual} introduce a modality-complementary prompter (MCP) block to fuse features from RGB and other modalities like thermal and depth.
\textbf{CMX}~\cite{zhang2023cmx} replicate the pretrained RGB encoder to tackle X modality, and place the proposed Feature Rectification Module (FRM) after each block to perform interaction of RGB features and X features. \textit{Note that these two baselines utilize the prior information about which channels are for RGB embedding while our framework does not utilize this information.}
We reimplement the above two methods on SAM following their original finetuning methods and evaluate their performance on our benchmark. 
As shown in Table~\ref{tab:compare_rgbx}, CMX~\cite{zhang2023cmx} does not achieve satisfying performance on finetuning the foundation model SAM. We suspect the unsatisfying performance is caused by the noise introduced from FRM, which appended after each block deviates the features from its original distribution, making the learning difficult. 
While ViPT~\cite{zhu2023visual} can achieve reasonable performance, its performance lags behind SimMAT.

\begin{figure}[h!]
  \centering
  \includegraphics[width=1\linewidth]{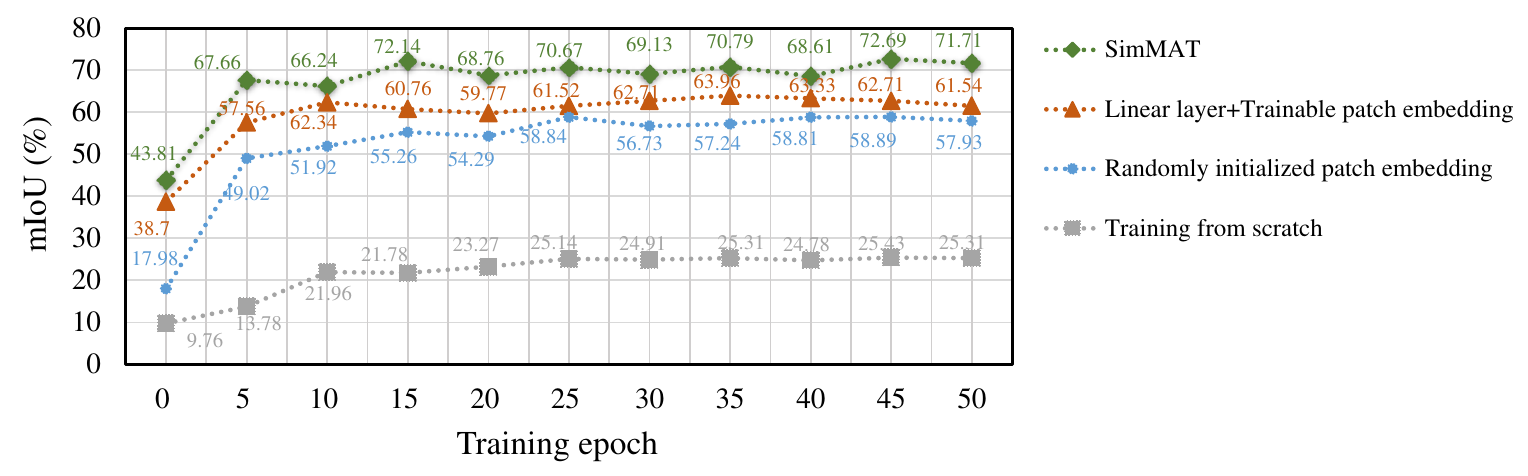}
  \caption{\textbf{The Training Curves for SimMAT and Baselines.} SimMAT achieves the best performance.}
  \label{fig:supp_curve}
\end{figure}

\begin{table*}[h!]
\small
\centering
\renewcommand{\arraystretch}{1.2}
\resizebox{0.95\linewidth}{!}{
\begin{tabular}{@{}l@{\hspace{3mm}}c@{\hspace{3mm}}l@{\hspace{3mm}}c@{\hspace{3mm}}c@{\hspace{3mm}}c@{\hspace{3mm}}c@{\hspace{3mm}}c@{\hspace{3mm}}c@{\hspace{3mm}}c@{\hspace{3mm}}c@{\hspace{3mm}}c@{}}
\toprule[1pt]

Method &  Params & Finetuning methods & RGB-T & RGB-D & RGB-HHA & RGB-NIR \\
    \midrule 
 CMX*~\cite{zhang2023cmx} & 403.8M & Full finetuning & 44.91 & 36.41 & 37.33 & 34.75 \\
 ViPT*~\cite{zhu2023visual} & 94.5M& Prompt tuning & 75.93 &48.89 &49.50 & 51.90 \\
\hline
SimMAT & \multirow{3}{*}{94.4M}  & LoRA  & 84.52 & 57.56 & 56.44 & \textbf{57.14} \\
SimMAT & & MLP Adapter &\textbf{85.29}  & \textbf{57.73}& \textbf{57.25} & 55.81 \\
SimMAT & & Full finetuning & 82.68 & 56.96 & 57.17 & 56.37 \\

\bottomrule[1pt]
\end{tabular}
}
\caption{\textbf{Comparison of SimMAT with Other Methods Tackling Pseudo New Modality (RGBX).} While with fewer parameters, SimMAT achieves better performance across four pseudo new modalities.  Note that ViPT and CMX can tackle RGBX only. * means reproduced implementation in SAM.
}
\label{tab:compare_rgbx}
\end{table*}

\clearpage

\section{Additional Details for Benchmark}
\label{sec:benchmark}
To study the problem of cross-modality transfer learning of SAM, we construct a new benchmark by collecting image segmentation datasets from different modalities, as described in the main paper. However, the segmentation labels of SAM are instance-level segmentation, but some segmentation datasets (\textit{e.g.,} ZJU-RGBP~\cite{zjurgbp}, NYUv2\cite{Silberman:ECCV12}) only provide semantic labels. Hence, {to align with the output of SAM}, we perform post-processing to convert the semantic labels to instance labels by decomposing non-connected components.

Figure~\ref{fig:supp_benchmark} shows the post-processing effect. Given a semantic map label, we partition it into separate masks if they are not pixel-connected to each other. Each separate mask serves as an instance label and is responsible only for the clicks that lie within it.
The evaluation metric IoU is calculated for each instance. Instead of average IoU over semantic categories, we take the average IoU of all instances as the mIoU results.

\begin{figure}[h]
  \centering
  \includegraphics[width=0.9\linewidth]{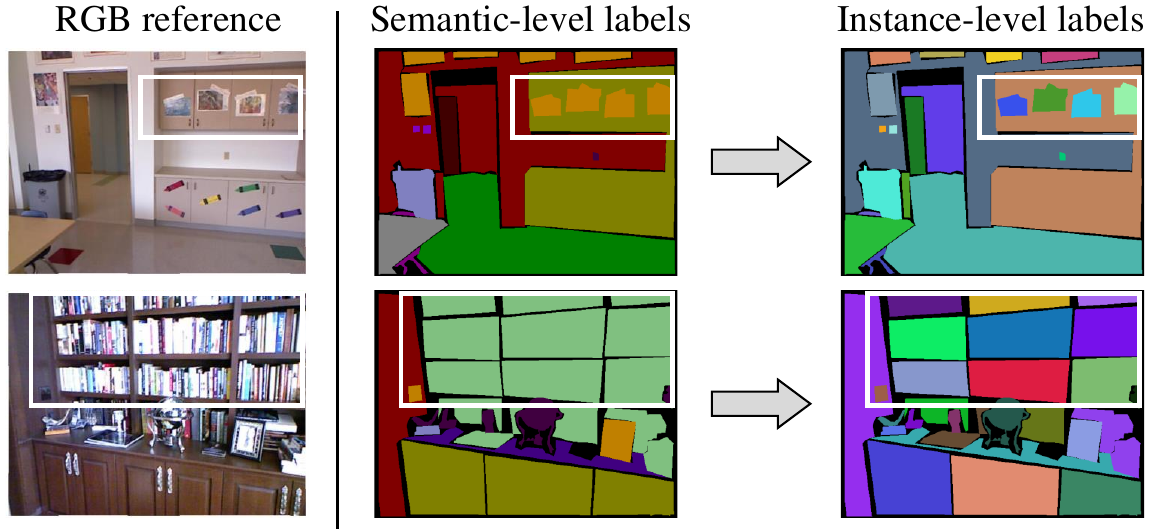}
  \caption{\textbf{The Illustration of Segmentation Generation Pipeline in Our Benchmark.} The semantic-level segmentation ground truth is split into instance-level segmentation ground truth.}
  \label{fig:supp_benchmark}
\end{figure}
\clearpage

\section{Additional Qualitative Results}
\label{sec:add_qualitative}
We provide more qualitative visualization results from Figure~\ref{fig:supp_depth} to Figure~\ref{fig:supp_hha}. For the SAM zero-shot performance, we use the provided RGB reference as the input. We present the results on diverse image modalities for better understanding. As shown in the figure, the performance of training from scratch and zero-shot is generally unsatisfying. {With our proposed SimMAT framework, the segmentation performance can be improved significantly.} 
For example, in the first column of thermal modality in Figure~\ref{fig:supp_thermal}, {we can see that both training from scratch and zero-shot fail to segment the ``window'' completely.} As a comparison, our method achieves accurate segmentation, which is quite close to the ground truth mask.

\begin{figure}
  \centering
  \includegraphics[width=0.9\linewidth]{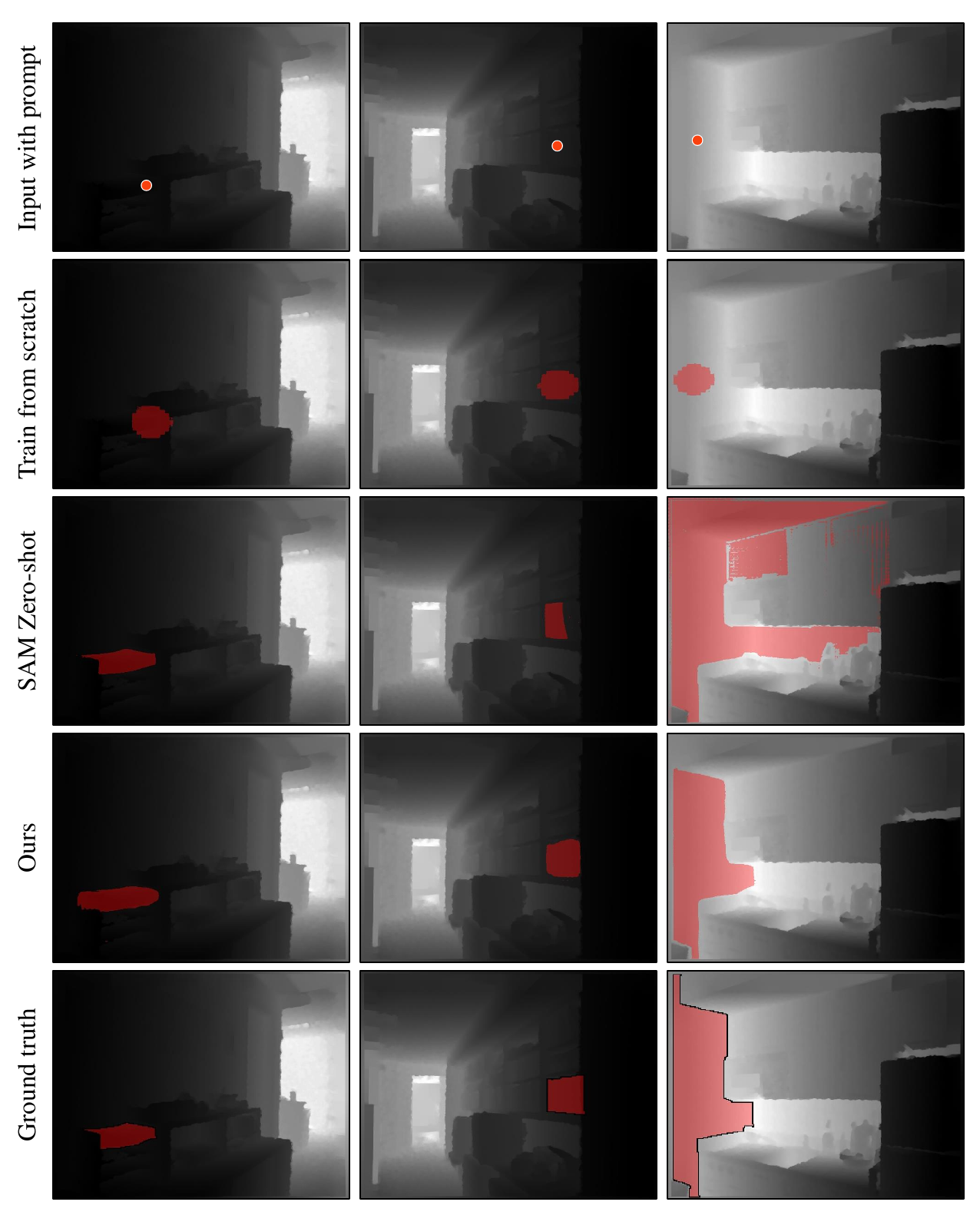}
  \caption{\textbf{Additional Qualitative Results in Depth Modality. } Our approach can perform better than zero-shot and training from scratch. }
  \label{fig:supp_depth}
\end{figure}

\begin{figure}
  \centering
  \includegraphics[width=0.9\linewidth]{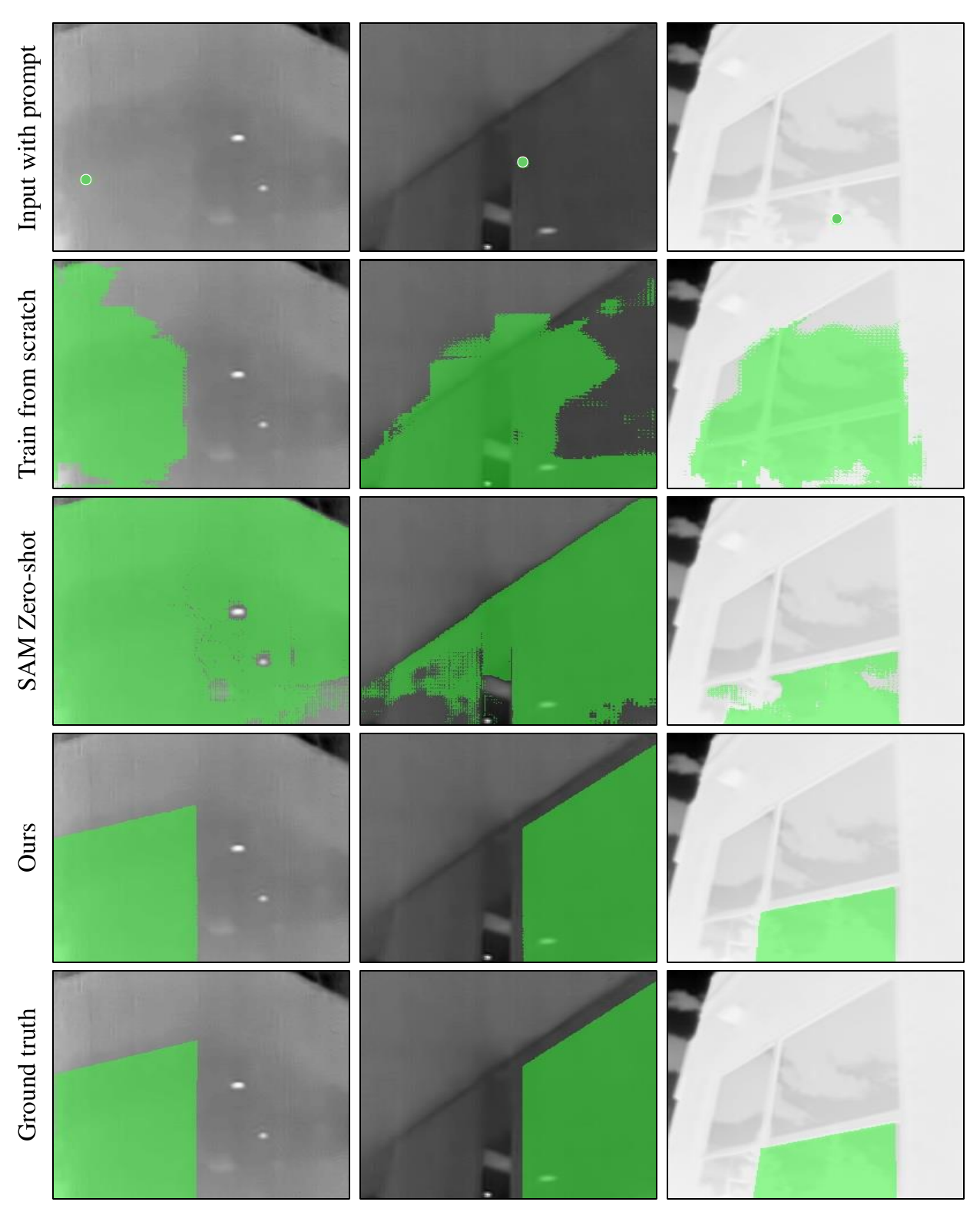}
  \caption{\textbf{Additional Qualitative Results in Thermal Modality. } Our approach can perform better than zero-shot and training from scratch. }
  \label{fig:supp_thermal}
\end{figure}

\begin{figure}
  \centering
  \includegraphics[width=0.9\linewidth]{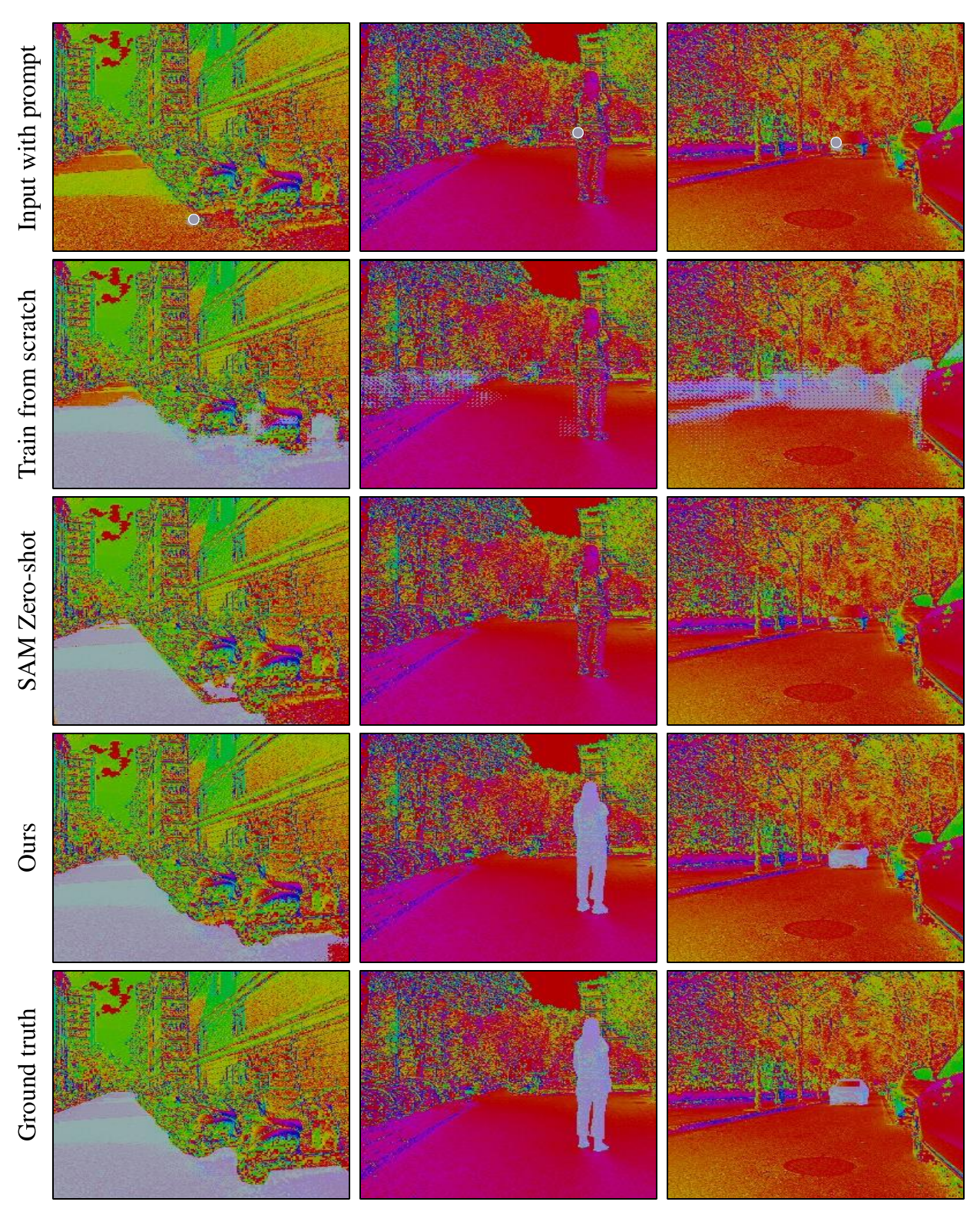}
  \caption{\textbf{Additional Qualitative Results in Polarization Modality. } Our approach can perform better than zero-shot and training from scratch. }
  \label{fig:supp_polarization}
\end{figure}

\begin{figure}
  \centering
  \includegraphics[width=0.9\linewidth]{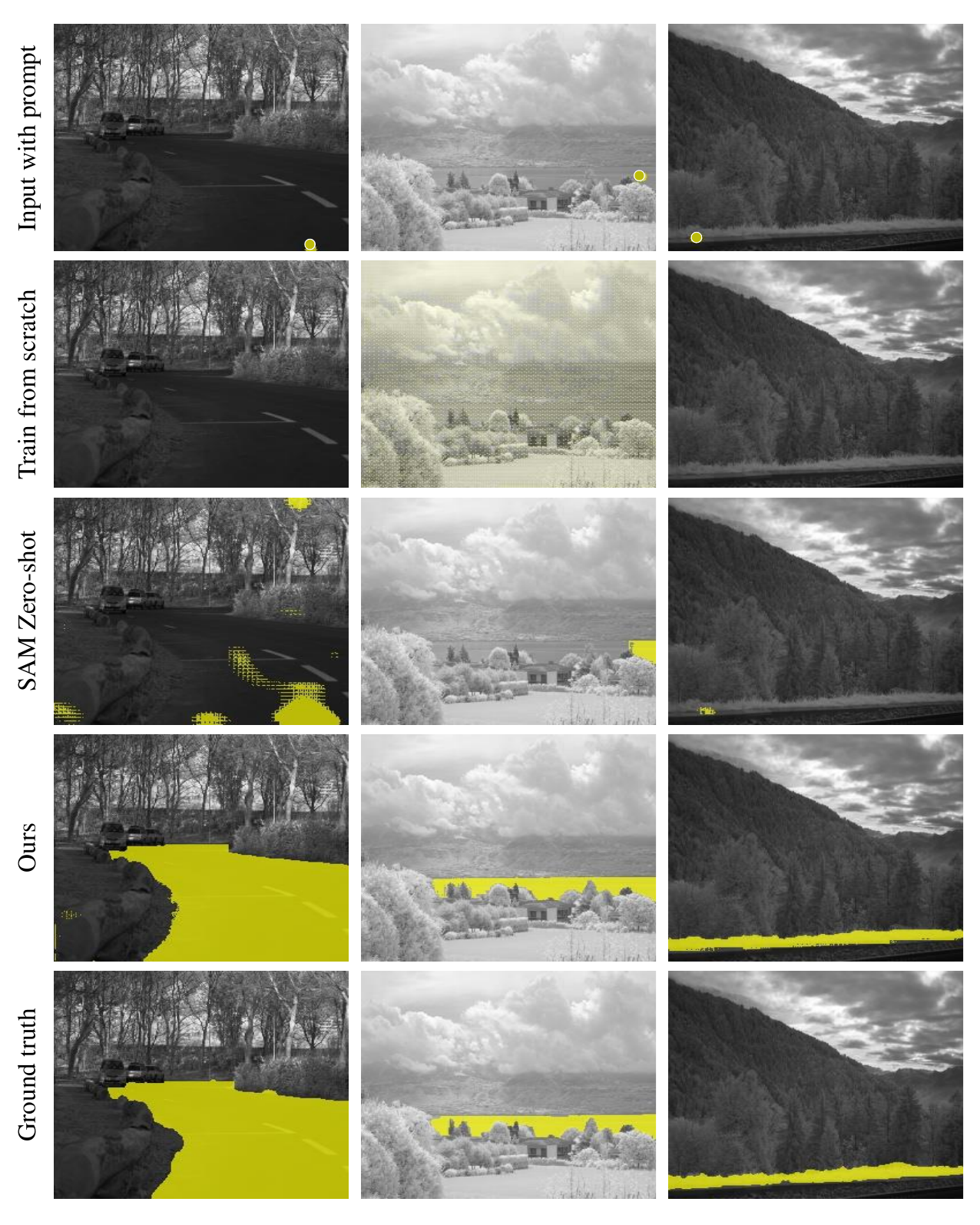}
  \caption{\textbf{Additional Qualitative Results in NIR Modality. } Our approach can perform better than zero-shot and training from scratch. }
  \label{fig:supp_nir}
\end{figure}

\begin{figure}
  \centering
  \includegraphics[width=0.9\linewidth]{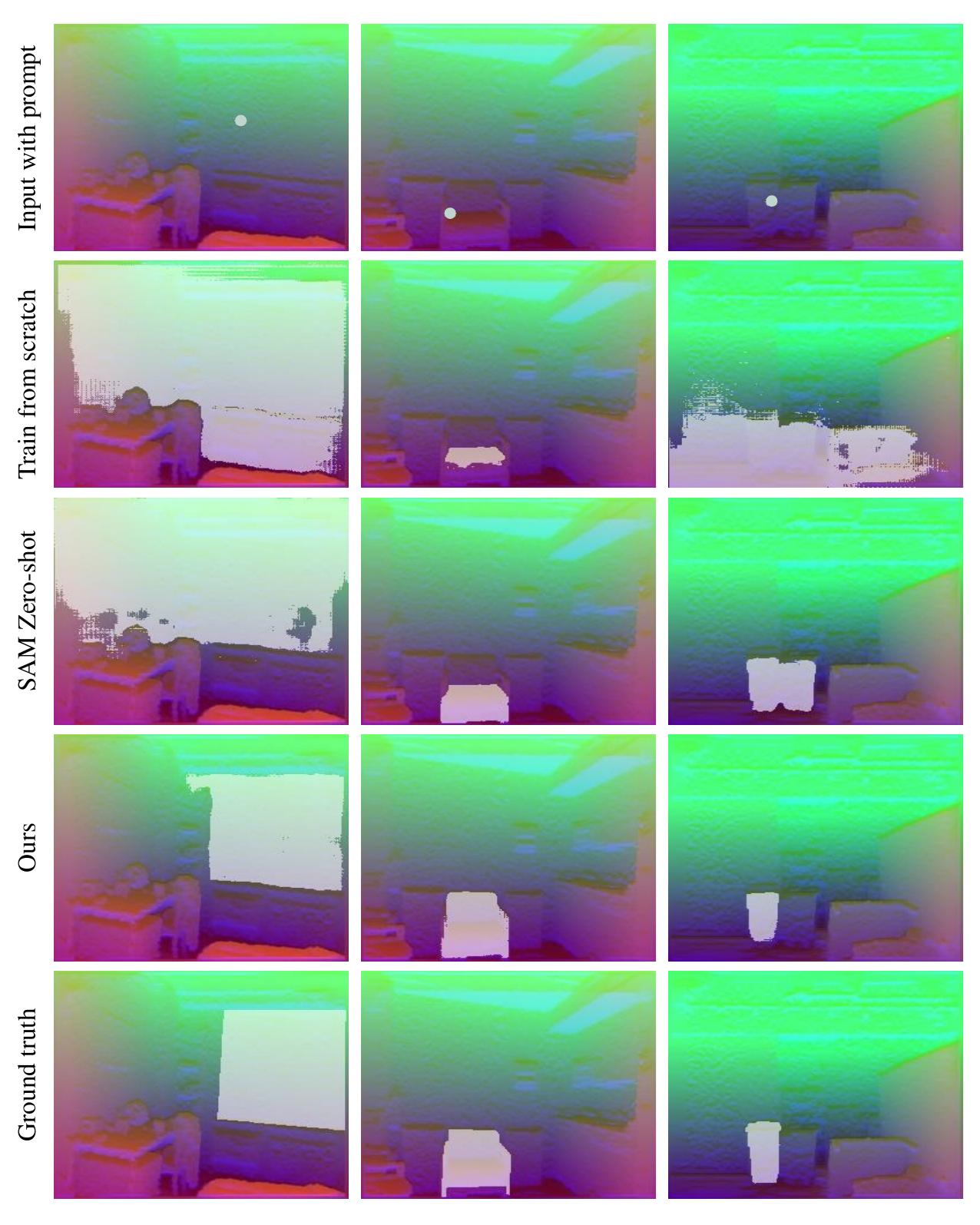}
  \caption{\textbf{Additional Qualitative Results in HHA Modality. } Our approach can perform better than zero-shot and training from scratch. }
  \label{fig:supp_hha}
\end{figure}

\end{document}